\documentclass[sigconf]{acmart}
\AtBeginDocument{%
  }

\setcopyright{acmcopyright}
\copyrightyear{2023}
\acmYear{2023}
\setcopyright{acmlicensed}\acmConference[KDD '23]{Proceedings of the 29th ACM SIGKDD Conference on Knowledge Discovery and Data Mining}{August 6--10, 2023}{Long Beach, CA, USA}
\acmBooktitle{Proceedings of the 29th ACM SIGKDD Conference on Knowledge Discovery and Data Mining (KDD '23), August 6--10, 2023, Long Beach, CA, USA}
\acmPrice{15.00}
\acmDOI{10.1145/3580305.3599848}
\acmISBN{979-8-4007-0103-0/23/08}
\usepackage{xcolor}
\usepackage{comment}
\usepackage{amsmath}
\usepackage{subfigure}
\usepackage{algorithm2e}
\newtheorem{remark}{Remark}
\usepackage{siunitx}

\SetKwInput{KwInput}{Input}
\SetKwInput{KwOutput}{Output}
\begin{document}

%%
%% The "title" command has an optional parameter,
%% allowing the author to define a "short title" to be used in page headers.
\title{Interactive Generalized Additive Model and Its Applications in Electric Load Forecasting}

\author{Linxiao Yang}
\email{linxiao.ylx@alibaba-inc.com}
 \affiliation{\institution{DAMO Academy, Alibaba Group}\city{Hangzhou}\country{China}}

\author{
Rui Ren}
\email{renrui.ren@alibaba-inc.com}
 \affiliation{\institution{DAMO Academy, Alibaba Group}\city{Hangzhou}\country{China}}

\author{
Xinyue Gu}
\email{guxinyue.gxy@alibaba-inc.com}
 \affiliation{\institution{DAMO Academy, Alibaba Group}\city{Hangzhou}\country{China}}

\author{
Liang Sun}
\email{liang.sun@alibaba-inc.com}
 \affiliation{\institution{DAMO Academy, Alibaba Group}\city{Hangzhou}\country{China}}

\renewcommand{\shortauthors}{Linxiao Yang, Rui Ren, Xinyue Gu, \& Liang Sun}
%% No italics

%%
%% The abstract is a short summary of the work to be presented in the
%% article.
\begin{abstract}
  % Electric load forecasting is an important building block for planning in power system industry. The growing cases of power outage are putting pressure on machine learning models for higher precision. The crux of the problem is to handle challenging circumstances when there are extreme weather events. In this paper, we allow human experts to incorporate domain knowledge to the model for performance improvement. We propose an interactive GAM that is human interpretable and human knowledge understandable. This boosting-based GAM leverages piecewise linear functions and can be learned through our efficient algorithm. In both public benchmark and electricity datasets, our interactive GAM achieves accuracy outperforming current state-of-the-art methods and demonstrates good generalization ability in the cases of extreme weather events. We also launch a user-friendly web-based product and put the method into real application.  

Electric load forecasting is an indispensable component of electric power system planning and management. Inaccurate load forecasting may lead to the threat of outages or a waste of energy. Accurate electric load forecasting is challenging when there is limited data or even no data, such as load forecasting in holiday, or under extreme weather conditions. As high-stakes decision-making usually follows after load forecasting, model interpretability is crucial for the adoption of forecasting models. In this paper, we propose an interactive GAM which is not only interpretable but also can incorporate specific domain knowledge in electric power industry for improved performance. This boosting-based GAM leverages piecewise linear functions and can be learned through our efficient algorithm. In both public benchmark and electricity datasets, our interactive GAM outperforms current state-of-the-art methods and demonstrates good generalization ability in the cases of extreme weather events. We launched a user-friendly web-based tool based on interactive GAM and already incorporated it into our eForecaster product, a unified AI
platform for electricity forecasting. 
\end{abstract}

%%
%% The code below is generated by the tool at http://dl.acm.org/ccs.cfm.
%% Please copy and paste the code instead of the example below.
%%


%%
%% Keywords. The author(s) should pick words that accurately describe
%% the work being presented. Separate the keywords with commas.
%% A "teaser" image appears between the author and affiliation
%% information and the body of the document, and typically spans the
%% page.

% \begin{teaserfigure}
%   \includegraphics[width=\textwidth]{sampleteaser}
%   \caption{Seattle Mariners at Spring Training, 2010.}
%   \Description{Enjoying the baseball game from the third-base
%   seats. Ichiro Suzuki preparing to bat.}
%   \label{fig:teaser}
% \end{teaserfigure}

% \received{20 February 2007}
% \received[revised]{12 March 2009}
% \received[accepted]{5 June 2009}

%%
%% This command processes the author and affiliation and title
%% information and builds the first part of the formatted document.
\maketitle
\section{Introduction}\label{sec:intro}

The electric load forecasting (ELF)~\cite{ELF:2017:survey} aims to predict future electricity consumption with historical observation data and other external factors such as numeric weather prediction (NWP). ELF is an indispensable procedure for the planning of power system industry~\cite{ELFbook:2021} to meet the demand and supply equilibrium, as the decision-making process counts heavily on load prediction. With extreme weather events becoming more common, however, it is challenging to model the electricity demand patterns and to provide accurate predictions. A prediction too conservative can result in the threat of blackouts, whereas an aggressive prediction will make surplus electricity accrue, which can be a big waste particularly for non-regeneration energy. For example, California Independent System Operator (CAISO) found out that under-scheduling of demand by around 4.5\% in the day-ahead market is the primary root cause for its proactive triggering of rotating blackouts in August 2020. A load forecast error of 1\% in terms of mean absolute percentage error (MAPE) can turn into several hundred thousand dollars expenditure per GW peak for a utility’s net income~\cite{hong2016probabilistic}, let alone the inconvenience brought to the society.

Accurate electric load forecasting is challenging when there is limited data, such as load forecasting in holiday, or under extreme weather conditions (e.g., high temperature, cold wave, typhoon). These scenarios rarely happen but potentially have serious consequences and therefore draw tremendous attention of the power industry and the public sector. This problem becomes even more challenging when there is no training data. For example, given a record breaking high temperature in summer, the electric load also increases significantly. As there is no historical data with such high temperature, many machine learning models, including the popular decision tree ensemble methods, fail to make accurate prediction in this case.

In the traditional electric power industry, generally human experts use the results predicted by ELF for further usage in different applications, e.g., capacity planning and network planning~\cite{ELFbook:2021}. As the human is in the loop, the interpretability of the forecasting model is crucial for the adoption of ELF models, especially when the prediction of ELF models and human experiences are inconsistent. Considering the nature of high-stakes decisions based on ELF, it is important to explain the model why it has reached a particular numerical result of predictions. 
% ELF used to be an experience-intensive procedure. Thus we suggest to let human expert knowledge be invoked to take account of these predicaments. This requires the model to be human interpretable. 
% Since the model will be applied into such a high risk and high-stakes area, it is crucial that the model can explain why it has reached a particular numerical result of predictions. 
Then experts are able to verify predictions so as to fine-tune models, troubleshoot or gain newer insights. Vice versa, human's domain knowledge should also be incorporated into forecasting models. Translating knowledge into mathematical language is necessary to integrate prior knowledge or human supervision into the learning process. As \cite{RudinEtAlSurvey2022} remarked that, interpretability allows human-machine collaboration and eventually leads to better performance. 

% {\color{red} key idea: model interpretability. Flow: Model interpretability importance and requirement in electric power industry--> incorporating domain knowledge into model leads to both  better model performance and model interpretability}

Among these interpretable machine learning models, the Generalized Additive Model (GAM) has been particularly successful for dealing with large datasets and elucidating complex relationships~\cite{RudinEtAlSurvey2022}. GAM is clearly interpretable because the model is a linear combination of univariate component functions. It reveals the contribution and the effect of a single factor to the prediction by plotting a curve of the target variable as a function of each component. Besides interpretability, GAM has two other significant benefits for ELF. In practice, it not only yields accuracy as competitive as other sophisticated machine learning models~\cite{wang2016electric,hong2019global}, but also fast to compute and therefore easy to deploy in industry. Thus, we choose to leverage GAM to develop our human-machine interactive ELF model in this study.

\begin{figure}
    \centering
    \vspace{-.3cm}
    \includegraphics[width=0.5\linewidth]{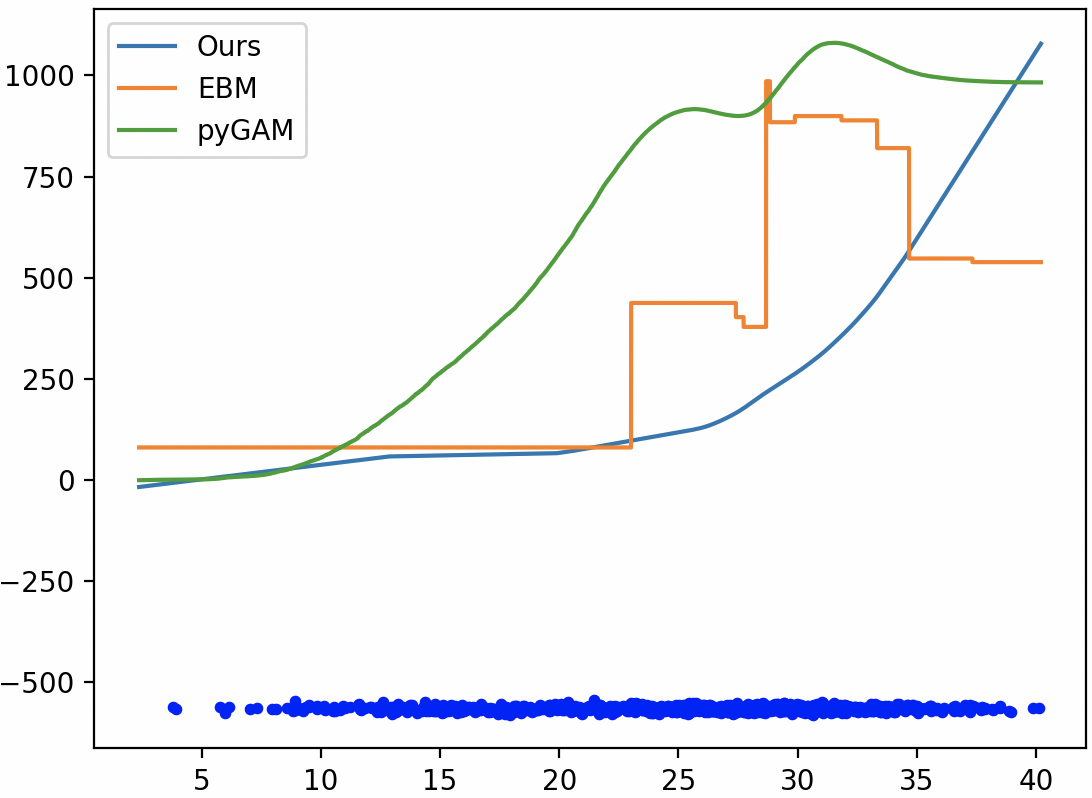}
    \vspace{-.3cm}
    \caption{Existing GAMs (green line and orange line) fail to extrapolate at rightmost scenarios when examples are limited, whereas our interactive GAM (blue line) finds the monotone pattern (see Section \ref{sec:exp_elf} for details).}
\vspace{-.5cm}
    \label{fig:drawbacks}
\end{figure}

There could be some drawbacks of GAM in the context of ELF. First, the challenge of tackling with extreme conditions still remains unsolved.
We conduct an experiment to demonstrate this phenomenon (detailed settings will be introduced in Section \ref{sec:exp_elf}). 
As the result in Figure~\ref{fig:drawbacks} shows, 
when the temperature climbs to an unprecedentedly high level (dots at bottom indicates frequency of such occurrences), two popular GAM algorithms, EBM~\cite{ebm:2019} and PyGAM~\cite{serven2018pygam}, are overwhelmed by the limited data, and make stable or even decreasing predictions. Second, it cannot effectively absorb specific domain knowledge in electric power industry. Recently, an editable GAM~\cite{wang2022interpretability}  has been proposed, but it cannot handle the specific form of domain knowledge in ELF. For example, usually the load increases monotonically as temperature increases in summer. By incorporating this domain knowledge, our proposed interactive GAM outperforms EBM and PyGAM significantly, as illustrated in Figure \ref{fig:drawbacks}. 
% our proposed interactive GAM is shown to be antidotal due to the incorporation of that domain knowledge. 

% {\color{red} Good properties and limitations of GAM in ELF
% GAM is a good choice in ELF. 1. explainable 2. competitive in precision, as some important factors impact the load significantly, and some GAM model proposed in the electric power industry~\cite{wang2016electric} . 3. fast and easy to deploy.  \\
% drawbacks of GAM: 1. Cannot make accurate forecasting given limited data or no data; (e.g., EBM-based GAM fails to make good extrapolation in extreme temperatures (post a graph). ) 2. Cannot effectively incorporate specific domain knowledge in electric power industry. 
% }

In this paper, we propose a novel GAM as a prescription  for ELF to circumvent the above challenges. Our contributions includes:
\begin{enumerate}
    \item A new GAM based on piecewise linear functions. We conduct numerical experiments of both benchmark regression tasks and ELF tasks. With its excellent performance in limited data settings, our method shows great extrapolation ability which improves the forecasting performance that users care about most.
    \item An interpretable ELF model. The proposed model can not only ``tell'' the key factors for electric load via visualization, but also ``hear'' the domain knowledge from experts in the electric power industry. We show how interpretability could lead to significant gains in ELF performance through the experiments.
    \item A user-friendly tool, which provides an  interactive web-based user interface for GAM. In particular, this tool has been incorporated into our eForecaster~\cite{eForecaster2023} product, a unified AI platform including robust, flexible, and explainable machine learning algorithms covering major electricity forecasting applications. Along with eForecaster, this tool has been deployed and applied in several provinces of China.  
\end{enumerate}

% {\color{red}
% Ours: propose an interactive GAM. contributions: \\
% 1. New GAM based on piecewise linear functions. \textbf{Novelty of methods} Good performance limited or no data; good extrapolation;  --> Improve forecasting performance the users care most; \\
% 2. Interpretable model, not only highlight the key factors but also incorporates domain knowledge from experts. For types of knowledge are defined and supported. Enables model understanding and improved forecasting performance. \textbf{Specific features} \\
% 3. A production deployed in real-world applications. 
% }
%reveals mechanisms how each factor affects the final result. 2. a production that is editable for users in power systems, which helps to refine the model and improve the performance. To make sure our minority class gets more importance in our predictions.
% 3. piece-wise linear, competitive accuracy. satisfy both accuracy and interpretability, and fast.

\section{Related Work}\label{sec:related:work}
In this section, we first review the current state-of-the-art algorithms for electric load forecasting (ELF), and then delve into the recent progress of GAMs with an emphasis of their applications in ELF. 

Electric load usually involves nonlinear modeling and exhibits seasonal patterns that change over time due to factors like weather changes~\cite{wang2016electric}, the economy drifts and holiday effect~\cite{dryar1944effect}. The electricity demand data, compared to sales data, can be of greater granularity temporally and spatially~\cite{hong2016probabilistic}. 
Based on the popular Global Energy Forecasting Competition (GEFCom) held in 2012, 2014 and 2017~\cite{hong2014global,hong2016probabilistic,hong2019global}, a series of competitions focusing on energy demand forecasting, the majority of the top-ranked solutions consider statistical machine learning methods~\cite{hastie2009elements}, including Gradient Boosting Machines (GBM) and Generalized Additive Models. 
%These methods have achieved competitive accuracy. 
In particular, the gradient boosting decision tree (GBDT)~\cite{hastie2009elements} and its variants, including XGBoost~\cite{chen2016xgboost}, CatBoost~\cite{prokhorenkova2018catboost}, LightGBM~\cite{ke2017lightgbm}), have enjoyed tremendous popularity in the load forecasting community~\cite{landry2016probabilistic,huang2016semi,smyl2019machine}. By converting the forecasting problem to a regression problem, GBDT can deal with seasonality as well as auto-correlation and its prediction achieves good performance most of the time. 
%By implementing aggregate algorithm over decision trees, GBDT gains explainability for the ensemble~\cite{delgado2022implementing}. 
However, it is reported that GBDT cannot predict well at the high loads~\cite{abbasi2019short}, possible due to the fact that high loads data points are under-represented and overlooked by the training process. 
Also, model extrapolation is tough for such tree-based ensemble learners~\cite{meyer2021predicting} as regression tree usually assigns a constant for a split feature space.
This feature makes it difficult to tackle with scenarios where extrapolation is required. 
% thus the out-of-the-box prediction may fail when the data contains features in which the extrapolation is not implicit.
To summarize, GBDT and its variants are good forecasting models most of the time, but they fail to handle forecasting under extreme weather conditions.

Due to the huge success of deep neural networks~\cite{GoodBengCour16}, forecasting algorithms based on neural networks have become more and more popular. In late 1990s, the Electric Power Research Institute sponsored a project to build a short-term load forecasting system based on neural networks~\cite{khotanzad1998annstlf}. The resulting products are still being used by many power companies today. Later, a large body of literature has developed and applied deep learning models including LSTM-RNN~\cite{he2017load}, RBM~\cite{li2017building}, CNN~\cite{zahid2019electricity} and more recently, transformer-based network~\cite{zhou2022fedformer,chen2022learning}. Despite the strong fitting power, deep learning architectures often suffer from overfitting due to multiple layers~\cite{shi2017deep} and not well-interpretable to human. In particular, deep learning models usually require a large amount of training data, which is especially challenging for periods with extreme weather conditions where limited or even no training data is available. 

Another popular method for electric load forecasting is the highly explainable Generalized Additive Model~\cite{hastie2017generalized}. 
It is also one of the most successful methods used in the top-ranked solutions in GEFCom2012, 2014 and 2017 competitions~\cite{nedellec2014gefcom2012,gaillard2016additive,sigauke2017forecasting}. 
In real-world application of electric power system, GAM has been widely adopted mainly due to its good practical performance, especially in tackling with extreme weather conditions and volatile, fluctuating load. 
For example, projects sponsored by Australian Energy Market Operators~\cite{fan2011short,fan2012forecasting} have developed a modified semi-parametric version of GAM. 
An extension of GAM with LASSO and tensor product interactions have also demonstrated its power in forecasting South Africa data~\cite{sigauke2017forecasting}, making it a useful tool for system operators in power utility companies. 

Compared to neural networks and gradient boosting algorithms, GAM is favored by the industry for the following reasons. First, it identifies and characterizes nonlinear regression effects~\cite{hastie2009elements} and is able to jump the accuracy hurdles. In certain scenarios in ELF, GAM outperforms other methods while retains simplicity~\cite{gaillard2016additive}. 
% Though some level of sophistication is necessary to achieve high accuracy, the industry may not need the most high accuracy anyway. Those aforementioned successful projects have found a good balance between simplicity and accuracy. 
Another attractive feature of GAM is its interpretability. 
GAM is simple enough for human experts to read each component of the model, recognize which the significant predictors are and how a GAM works~\cite{hegselmann2020evaluation}. 
Thus, the effect of each factor on the prediction can be visualized in a graph, based on which  humans can work on the follow-up feedback, such as model-debiasing~\cite{tan2018distill}, or potentially, refining the under-fitting parts.

GAM can be trained using different types of components, such as splines~\cite{wood2003thin}, regression trees and tree ensembles~\cite{bauer1999empirical}. Splines are inefficient in modeling higher order interactions as number of parameters explodes exponentially~\cite{lou2013accurate}. Tree-based methods fit the curve in a piecewise constant way. Fitting GAM often involves backfitting~\cite{hastie2017generalized} or gradient boosting~\cite{friedman2001greedy}. Recently, some variants of GAM are proposed for improved accuracy~\cite{lou2012intelligible,petersen2019data,lou2013accurate,petersen2016fused}, such as GA$^2$M~\cite{lou2013accurate}, FLAM~\cite{petersen2016fused}. However, these GAMs cannot handle electric load forecasting when data is limited, especially in scenarios under extreme weather conditions. 
GAM Changer~\cite{wang2022interpretability}, an editable GAM tool is proposed to help domain experts and data scientists to fix problematic patterns. Unfortunately, it cannot deal with specific patterns of electric load effectively. In particular, when one function is modified, other functions still remain the same, potentially magnifying the fitting error. In other words, it lacks flexibility to handle complicated scenarios in real-world applications. 
% GAM Changer: when a function is modified, other functions remain the same.
% GAMs performance analysis 
%tree-based:EBM~\cite{nori2019interpretml} 
%spline:
% FLAM can perform poorly if the true conditional association between a feature and the response is truly linear~\cite{petersen2019data}
% application: healthcare \cite{caruana2015intelligible}
% fast algorithm interaction GA2M\cite{lou2013accurate}2. application: a. general, FLAM \cite{petersen2016fused} b: 

\vspace{-.2cm}
\section{Problem Formulation}\label{sec:iii}
%{\color{red}This paragraph can be moved to the introduction section}

%We consider the electric load forecasting under extreme whether event.
%As the extreme whether events happen rarely in the history, few data about the electric load can be collected.
%Fortunately, some domain knowledge from the experts can alleviate the lack of data about the electric load under extreme whether event.

%{\color{red}end of this part}

In ELF, the domain knowledge is usually regarded as ``trend knowledge" or ``curve shape knowledge". For example, when the temperature increases from $\SI{30}{\celsius}$ to $\SI{41}{\celsius}$, the urban electric load in summer also increases. 
The trend may also have increasing marginal effect. That is, $\SI{1}{\celsius}$ up leads to much more load increase at $\SI{36}{\celsius}$ than that 
at $\SI{34}{\celsius}$.
But when the temperature is too high (say above $\SI{41}{\celsius}$, basically all air conditioners are already running), increasing temperature only leads to fewer load rising.
Thus we can conclude that when the temperature is above about $\SI{27}{\celsius}$ and below about $\SI{41}{\celsius}$, the load is convex with respect to temperature and when the temperature is above about $\SI{41}{\celsius}$ the electric load is concave with respect to temperature.
%  And also the increase of electric load also increases as temperature increases in this range. 
However, these experts generally can determine the curve shape of the electric load as some key factors change (e.g., temperature), but they cannot quantify exact load change given these factors in general. 

%Formally, these domain knowledge have two key properties. The first type of knowledge is the so called the ``trend knowledge". Basically, we can only know the trend of load with respect to some factors, but cannot quantify the exact load value given these factors. The second type of knowledge 
%{\color{red} Secondly, even the high-level knowledge, one can not exactly determinate the range of the knowledge applied, such as even experienced experts can not exactly tell on which temperature level the load saturation happens.  --> This part should be rephrased }

Our interactive GAM translates domain knowledge into constraints. 
Specifically, let $\boldsymbol{x}_i\in\mathbb{R}^D$ and $y_i\in\mathbb{R}$ be a collection of $D$ features and the electric load at time index $i$, respectively. Usually, $\boldsymbol{x}_i$ includes weather prediction, time and electric load in history, etc. 
In GAMs, we formulate $y_i$ as the summation of the individual effects of $\{x_{i}^d\}_{d=1}^D$,
i.e.,
\begin{align}
y_i \approx \sum_{d=1}^D f_d(x_i^d),
\end{align}
where $x_{i}^d$ is the $d$th feature of $\boldsymbol{x}_i$, $f_d(\cdot)$ 
is the shape function of $x_{i}^d$.

To incorporate the domain knowledge, given a dataset $\{(\boldsymbol{x}_i,y_i)\}_{i=1}^N$, we aim to solve the following problem by introducing constraints: 
\begin{align}
\min_{\{f_d\}}&\;\sum_{i=1}^{N} w_i\left(y_i-\sum_{d=1}^D f_d(x_i^d)\right)^2\quad
\text{s.t.}\; f_d\in \mathcal{A}_d, d=1,2\dots D.
\end{align}
Here $w_i$ denotes the weight of $i$th sample, $\mathcal{A}_d$ denotes the set consists of all shape functions satisfying the constraints added by user.
We note that although we write the constraints in the formulation stage, actually, these constraints are added in an interactive manner. 
In other words, the constraints are added after the model is trained, and once the constraints added, the model is adjusted to fit these constraints.

We discuss ${\mathcal{A}_d}$, the feasible set of constraints for the $d$th feature. 
After extensive discussion with experts in the electric power industry, we define four types of constraints to represent the domain knowledge:
\begin{itemize}
    \item The monotone increasing constraint in some range;
    \item The monotone decreasing constraint at some range;
    \item Convex constraint in some range;
    \item Concave constraint in some range. 
\end{itemize}
For example, the monotone increasing constraints ensure that the electric load is monotone increasing when one of several factors increases in some range. 
% We perform a survey among the experts on electric power companies, and find that most domain knowledge can be classified as four types, i.e., the electric load is monotone increase, monotone decrease, convex, and concave with respective to some factors at some ranges.
% Thus we only consider these four types constraints in this paper. 
Note that the convex combination of two monotone increasing shape functions is also a monotone increasing function. This property also holds for monotone decreasing, convex, and concave constraints. 
Thus, ${\mathcal{A}_d}$ is a convex set based on its definition. 
Later we will show that this critical property empowers us to develop efficient algorithm to learn a set of shape functions to satisfy these constraints.

As discussed in Section \ref{sec:intro}, a key challenge in electric load forecasting under extreme weather conditions is the lack of training data in the past. For example, how to forecast the electric load at 40$^\circ$C in a city given the highest degree in the history collected is 38$^\circ$C. Thus, good model generalization on out-of-range data is crucial. 
We observe that for the electric load forecasting problem, the trend of the affection of factors (e.g., temperature) is usually smooth. 
For instance, the raising of the temperature from 38$^\circ$C to 39$^\circ$C leads to 10MW electric load increasing in a city recently, then we can also assume that the increase of electric load when the temperature increases from 39$^\circ$C to 40$^\circ$C is also around 10MW given other factors remaining the same. Thus, by discussing with domain experts and analyzing the historical data, we propose to restrict the shape function $f_d(\cdot)$ being piecewise linear. 

% When extreme whether events happens, some whether features are out of the range of its history values. 
% Thus, good generalization on out of range data is desired. 
% Usually, for the electric load forecasting problem, the trend of the affect of factors is smooth. 
% For instance, if raising the temperature from 38 to 39 Celsius degree leads to 10MW electric load increasing, then raising the temperature from 39 to 40 Celsius degree, electric load increasing is also around 10MW.
% To utilize this property, we further restrict the shape function $f_i(\cdot)$ being piecewise linear.

\section{Interactive GAM}
This section discusses our proposed interactive GAM. 
We first introduce the proposed piecewise linear GAM and then describe how we enable the model to be interactive with end users.

\subsection{Piecewise linear GAM}
\subsubsection{Overall Framework}
Our piecewise linear GAM is learned in the boosting framework. 
Specifically, we update the shape function $f_d(\cdot)$ according to the fitting residual in each iteration. At the $t$th iteration, fitting residual $\{r^{(t)}_i\}_{i=1}^N$ is 
\begin{align}
    r_i^{(t)} = y_i - \sum_{i=1}^N f_d^{(t-1)}(x_i^d), \label{eq:residual-update}
\end{align}
where $f_d^{(t-1)}$ denotes the estimate of $f_d$ at the $(t-1)$th iteration. Here we make an innocuous assumption that the mapping from $\{x^d_i\}_{i=1}^N$ to $\{r_i^{(t)}\}_{i=1}^N$ can be approximated using a piecewise linear function $g$, i.e.,
\begin{align}
    g = \textbf{PLA}(\{x^d_i\}_{i=1}^N,\{r_i^{(t)}\}_{i=1}^N,\{w_i\}_{i=1}^N), \label{eq:pla}
\end{align}
where $\textbf{PLA}(x,r,w)$ denotes the operator which approximates $r$ using a piecewise linear function of $x$, and $w$ denotes the weight of each sample.
%We will give more details on $\textbf{PLA}(x,y,w)$ in next subsection.

In the boosting framework, at the $t$th iteration, we update $f_d^{(t)}(x)$ with
\begin{align}
    f_d^{(t)}(x) = f_d^{(t-1)}(x) + \mu g(x),\label{eq:function-update}
\end{align}
where $\mu>0$ is the step size.
As we iteratively add piecewise linear functions to $f_d$, we then can ensure the final output $f_d$ to be a piecewise linear function.

We summarize our piecewise linear GAM in Algorithm \ref{alg:gam}. 

\RestyleAlgo{ruled}

%% This is needed if you want to add comments in
%% your algorithm with \Comment
\SetKwComment{Comment}{/* }{ */}

\begin{algorithm}
\caption{Piecewise Linear GAM}\label{alg:gam}
\KwInput{Training dataset $\{\boldsymbol{x}_i,y_i\}_{i=1}^N$, sample weight $\{w_i\}_{i=1}^N$, stepsize $\mu$, maximum iteration $T$}
\KwOutput{Shape function $\{f_d\}_{d=1}^D$}
 Initialization: $f_d^{(0)} = 0 \quad\forall d=1\dots D$\;
 \For{$t\gets1$ \KwTo $T$}{
 \For{$d\gets1$ \KwTo $D$}{
 Step 1: compute $\{r_i^{(t)}\}$ according to (\ref{eq:residual-update})\\
 Step 2: approximate the relationship between $\{x_i^d\}$ and $\{r_i^{(t)}\}$ according to (\ref{eq:pla})\\
 Step 3: update shape function $f_d^{(t)}$
according to (\ref{eq:function-update})
 }
 }
\end{algorithm}
% In this section, we propose a user-customized approach to fit an additive model, which is estimated to be piecewise linear with a group of basis functions. Compared with piecewise constant, our piecewise linear method could have better performance when the data sample is sparse.
\vspace{-.3cm}
% \emph{Discussion:} 
\begin{remark}
We ensure the smoothness of $f_i$ by forcing $g(x)$ to be smooth, instead of forcing $f_i$ as backfitting based methods do. 
After several rounds of boosting, one can still get an $f_i$ with edges.
Compared with backfitting based methods, our method is more flexible.
\end{remark}

\subsubsection{Piecewise linear approximation}
We now introduce our piecewise linear approximation method.
Given a dataset $\{(x_i,r_i,w_i)\}_{i=1}^{N}$, we approximate $r_i$ via a piecewise linear function of $x_i$ with weight $w_i$. This is equivalent to the following optimization problem: 
\begin{align}
    \min_{g\in \mathcal{L}} \sum_{i=1}^N w_i(g(x_i)-r_i)^2,
\end{align}
where $\mathcal{L}$ denotes the set of all possible piecewise
linear functions.
The problem of approximating data points via
piecewise linear function has been studied for
years~\cite{magnani2009convex,d2010piecewise,siahkamari2020piecewise}. 
In this paper, we assume $g$ is a weighted combination of 
hinge functions, reverse hinge functions and a constant:
\begin{align}
    g(x) = c + \sum_l \phi_l h(x-\eta_l) + \sum_l \phi_l^r h^r(x-\eta_l),\label{def:g_x}
\end{align}
where $c$ is an unknown constant, $\{\eta_l\}$ denotes the respect parameters, and $\{\phi_l\}$ and $\{\phi_l^r\}$ denote the weight of each functions. The hinge function $h(x)$ and reverse hinge function $h^r(x)$ are defined as $\max(x,0)$ and $-\min(x,0)$, respectively. 
Note that the hinge function is already used to learn piecewise linear functions in the literature. For example, in multivariate adaptive regression splines (MARS) models~\cite{hastie2009elements}, hinge function is used as the basis function.

To find the optimal $\{c, \eta, \phi, \phi^r\}$, we first restrict $\{\eta_l\}$ in a known finite set of size $L$. 
% Let $\eta_l$ be the $l$th possible value $\eta$ can take, 
We form matrix $\boldsymbol{A}^h\in\mathbb{R}^{N\times L}$, whose 
$(i,l)$th elements is $h(x_i-\eta_l)$, and matrix $\boldsymbol{A}^r\in\mathbb{R}^{N\times L}$, whose 
$(i,l)$th elements is $h^r(x_i-\eta_l)$.
Stacking $\{g(x_i)\}$ as a vector, we have
$[g(x_1),\dots,g(x_N)]^T=\boldsymbol{A}\boldsymbol{q}$,
where 
\begin{align}
    \boldsymbol{A}=[\boldsymbol{1},\boldsymbol{A}^h,\boldsymbol{A}^r],\label{omp:def-a}
\end{align}
$\boldsymbol{1}\in\mathbb{R}^{N\times 1}$ is a vector whose all elements is one, and 
\begin{align}
    \boldsymbol{q}=[c,\phi_1,\dots,\phi_L,\phi^r_1,\dots,\phi^r_L]^T
\end{align}
denotes the unknown parameters. Formally, finding the optimal piecewise linear approximation of $\{r_i\}$ can be formulated as the following optimization problem: 
\begin{align}
\min_{\boldsymbol{q}}\; \|\boldsymbol{W}^{\frac{1}{2}}(\boldsymbol{r}-\boldsymbol{A}\boldsymbol{q})\|_2^2,\qquad
\text{s.t.}  \; \|\boldsymbol{q}\|_0\le K, \label{problem:omp-ori}
\end{align}
where $\boldsymbol{W}\in\mathcal{R}^{N\times N}$ denotes a diagonal matrix whose $(i,i)$th entry equals to $w_i$, $\|\cdot\|_0$ is the $\ell_0$ pseudo-norm that counts the number of nonzero entries in the vector, and the constraint in (\ref{problem:omp-ori}) restricts $g(x)$ be formed by at most $K$ $(K\ll N)$ constant, hinge
or reverse hinge functions. We note that this constraint can significantly reduce the computational complexity.

To ensure the generalization of the proposed method, instead of directly solving
the problem (\ref{problem:omp-ori}), we solve the following regularized problem:
\begin{align}
\min_{\boldsymbol{q}}\; \frac{1}{N}\|\boldsymbol{W}^{\frac{1}{2}}(\boldsymbol{r}-\boldsymbol{A}\boldsymbol{q})\|_2^2+\lambda \|\boldsymbol{q}\|_2^2,\qquad
\text{s.t.} & \; \|\boldsymbol{q}\|_0\le K,\label{problem:omp-new}
\end{align}
where we add a $\ell_2$-norm regularization on $\boldsymbol{q}$ to avoid
the shape change of $g(x)$, and parameter $\lambda$ controls the trade-off between the fitting error and $\ell_2$-norm of $\boldsymbol{q}$.

Note that problem (\ref{problem:omp-new}) is a variant of the sparse regression problem. The sparse regression problem has motivated a flourishing line of work~\cite{zhang2015survey,wen2018survey,wright2010sparse}.
Greedy methods, such as orthogonal matching pursuit (OMP), are one of the most popular methods, due to their computational efficiency. 
In this paper, we extend OMP~\cite{omp} to solve (\ref{problem:omp-new}). 
As the vector $\boldsymbol{q}$ is sparse, once we can estimate the positions of nonzero elements of $\boldsymbol{q}$, we can find the optimal $\boldsymbol{q}$ by solving a least squares problem.
Here we find the nonzero positions of $\boldsymbol{q}$ by their corresponding columns of matrix $\boldsymbol{A}$.
Specifically, we maintain a set $\Gamma$ (initially empty) and keep adding to $\Gamma$ with the column index of $\boldsymbol{A}$ that most reduces the objective value until the size of $\Gamma$ equals to $K$.
Define $\boldsymbol{A}_{\Gamma}$ be a matrix that formed by set $\{\boldsymbol{a}_i|i\in \Gamma\}$, where $\boldsymbol{a}_i$ denotes the $i$th columns of $\boldsymbol{A}$. At the $k$th iteration, let $\boldsymbol{q}_{\Gamma}$ be the weight vector estimate with respect to $\boldsymbol{A}_{\Gamma}$. 
The fitting residual $\boldsymbol{b}$ is
$$
\boldsymbol{b}=\boldsymbol{r}-\boldsymbol{A}_{\Gamma}\boldsymbol{q}'.
$$
Let $(\boldsymbol{a},q)$ be the pair of column of $\boldsymbol{A}$ and its corresponding weight.
At the $k$th iteration, we aim to find a pair of $(\boldsymbol{a},q)$ minimizing
\begin{align}
g(\boldsymbol{a},q)= \frac{1}{N}\|\boldsymbol{W}^{\frac{1}{2}}(\boldsymbol{b}-q\boldsymbol{a})\|_2^2+\lambda q^2.\label{obj:omp}
\end{align}
Fixing $\boldsymbol{a}$, we can find optimal $q$ that miniming (\ref{obj:omp}) is given as 
\begin{align}
q^{*}= \frac{\boldsymbol{b}^T\boldsymbol{W}\boldsymbol{a}}{\boldsymbol{a}^T\boldsymbol{W}\boldsymbol{a}+\lambda N}. 
\end{align}
Substituting it into (\ref{obj:omp}), we arrive at
\begin{align}
g(\boldsymbol{a},q^{*})= \frac{1}{N}\boldsymbol{b}^T\boldsymbol{W}\boldsymbol{b}-\frac{\boldsymbol{b}^T\boldsymbol{W}\boldsymbol{a}}{\boldsymbol{a}^T\boldsymbol{W}\boldsymbol{a}+\lambda N}. \label{omp:selector}
\end{align}
Then at the $k$th iteration, we only need to find an $\boldsymbol{a}$ minimizing (\ref{omp:selector}).
Once the optimal $\boldsymbol{a}$ is found, its index is added into
$\Gamma$, and $\boldsymbol{A}_{\Gamma}$ is updated as 
$\boldsymbol{A}_{\Gamma}=[\boldsymbol{A}_{\Gamma},\boldsymbol{a}]$.
Then the optimal weight $\boldsymbol{q}_{\Gamma}$ can be updated
by minimizing
\begin{align}
\min_{\boldsymbol{q}_{\Gamma}}&\quad \frac{1}{N}\|\boldsymbol{W}^{\frac{1}{2}}(\boldsymbol{r}-\boldsymbol{A}_{\Gamma}\boldsymbol{q}_{\Gamma})\|_2^2+\lambda \|\boldsymbol{q}_{\Gamma}\|_2^2, 
\end{align}
which is a least squares problem and its optimal solution is 
given as follows
\begin{align}
\boldsymbol{q}^*_{\Gamma} = (\boldsymbol{A}_{\Gamma}^T\boldsymbol{W}\boldsymbol{A}_{\Gamma}+\lambda N\boldsymbol{I})^{-1}\boldsymbol{A}_{\Gamma}^T\boldsymbol{W}\boldsymbol{r}, \label{omp:q-update}
\end{align}
where $\boldsymbol{I}$ denotes the identity
matrix.
Once $K$ pairs of $(\boldsymbol{a},p)$ are estimated, 
we estimate
$\boldsymbol{q}$ by replacing the elements at
positions in $\Gamma$ of a zero vector by $\boldsymbol{q}_{\Gamma}^*$.

We summarize our piecewise linear approximation method
in Algorithm~\ref{alg:omp}.

\RestyleAlgo{ruled}

%% This is needed if you want to add comments in
%% your algorithm with \Comment
\SetKwComment{Comment}{/* }{ */}

\begin{algorithm}
\caption{the proposed Piecewise Linear Algorithm (PLA)}\label{alg:omp}
\KwInput{Dataset $\{(x_i,r_i,w_i)\}_{i=1}^N$, threshold set $\{\eta_l\}_{l=1}^L$, trade-off parameter $\lambda$, maximum number of basis $K$.}
\KwOutput{$g(x)$}
 Initialization: form matrix $\boldsymbol{A}$ according to (\ref{omp:def-a}), $\boldsymbol{b}=y$, and $\Gamma=\emptyset$;\\
 \For{$k\gets1$ \KwTo $K$}{
 Step 1: $\gamma_{k}=\mathop{\arg\max}\limits_{j \notin \Gamma}\frac{\boldsymbol{b}^T\boldsymbol{W}\boldsymbol{a}_j}{\boldsymbol{a}_j^T\boldsymbol{W}\boldsymbol{a}_j+\lambda N}$\\
 Step 2: $\Gamma=\Gamma \cup \gamma_{k}$ and $\boldsymbol{A}_{\Gamma}=[\boldsymbol{A}_{\Gamma},\boldsymbol{a}_{\gamma_k}]$\\
 Step 3: compute $\boldsymbol{q}_{\Gamma}^*$ according to (\ref{omp:q-update})\\
 Step 4: $\boldsymbol{b}=\boldsymbol{r}-\boldsymbol{A}_{\Gamma}\boldsymbol{q}_{\Gamma}^*$\\
 }
 Step 5: 
Generate $g(x)$ according to (\ref{def:g_x}) with $\boldsymbol{q}_{\Gamma}^*$\\
\end{algorithm}
\vspace{-.5cm}
\begin{remark} 
For better generalization, we add a $\ell_2$-norm regularization on $\boldsymbol{p}$, which penalizes the slope of the hinge functions.
$\lambda$ controls the trade-off
between smoothness and fitting error. Large $\lambda$ leads to small
$\boldsymbol{q}$ and more smooth $f_d(x)$. 
To reduce the search space, we use one $\lambda$ for all the features.
$\boldsymbol{q}$ is also affected by the Frobenius norm of $\boldsymbol{A}$.
To wipe the effect of the value ranges of features on the magnitude of $\boldsymbol{A}$, we normalize the features before training.
Apart from regularization terms, we can regularize the objective by pairwise selection for hinge function and reverse hinge function. 
In this way, $\gamma_k$ is determined by
\begin{align}
    \gamma_{k}=\mathop{\arg\max}\limits_{j \notin \Gamma}\frac{\boldsymbol{b}^T\boldsymbol{W}\boldsymbol{a}^h_j}{(\boldsymbol{a}^h_j)^T\boldsymbol{W}\boldsymbol{a}^h_j+\lambda N}+\frac{\boldsymbol{b}^T\boldsymbol{W}\boldsymbol{a}^r_j}{(\boldsymbol{a}^r_j)^T\boldsymbol{W}\boldsymbol{a}^r_j+\lambda N},
\end{align}
where $\boldsymbol{a}^h_j$ and $\boldsymbol{a}^r_j$ denote the $j$th column of $\boldsymbol{A}^h$ and $\boldsymbol{A}^r$, respectively.
The pairwise selection avoids only fitting a small number of data, which leads to model overfitting. 
% We that such a strategy that pairwise adding functions into consideration is also used in MARS. 
\end{remark}

\begin{remark}
We discuss the computational complexity of the proposed method. The main computational complexity of the proposed method is dominated by Step 1 and Step 3 in Algorithm \ref{alg:omp}. 
Step 1 evaluates all the possible thresholds with computational complexity $\mathcal{O}(NL)$. 
For Step 3, although inverse operation is involved, $\boldsymbol{q}_{\Gamma}^*$ can be estimated efficiently by solving following equation 
\begin{equation}
(\boldsymbol{A}_{\Gamma}^T\boldsymbol{W}\boldsymbol{A}_{\Gamma}+\lambda N \boldsymbol{I})\boldsymbol{q}^*_{\Gamma} = \boldsymbol{A}_{\Gamma}^T\boldsymbol{W}\boldsymbol{r},
\end{equation}
with computational complexity $\mathcal{O}(K^2)$. 
In our experiments, $K$ is set to a quite small number, e.g., 5. 
Thus,  the main computational cost lies in Step 1. 
Since the more thresholds are considered, the more accurate the approximation is, we tend to select large $L$, which further increases the computational complexity.
To reduce the computational complexity, in the next section, we propose an efficient algorithm which reduces the computational complexity of Step 1 to $\mathcal{O}(N+L)$.
\end{remark}

\subsection{Efficient Algorithm}
In this subsection we introduce an efficient algoithm to speed up the computation in Step 1 of Algorithm \ref{alg:omp} by utilizing the special structure of the hinge function. 
% As discuss above, the computation complexity of step 1 in Alg.\ref{alg:omp} is main bottleneck of the proposed method. 
%Fortunately, by utilizing the special structure of hinge function, the computation complexity of step 1 in Alg.\ref{alg:omp} is reduced to be linear with $N$ and $L$.
Without loss of generality, we assume the dataset $\{(x_i,r_i,w_i)\}_{i=1}^N$ is sorted according to $x_i$
such that $x_1\le x_2\le\dots\le x_N$,
$\{\eta_l\}$ are also sorted such as 
$\eta_1\le \eta_2\le\dots\le\eta_L$,
and $\eta_l$ corresponding to $\boldsymbol{a}_l^h$ $($and $\boldsymbol{a}_l^r$ $)$. 
For the $l$th threshold $\eta_l$, let $l' (1\le l'\le N)$ be the index such that
$x_{l'-1}<\eta_l\le x_{l'}$.
Then $(\boldsymbol{a}_l^h)^T\boldsymbol{W}\boldsymbol{a}_l^h$ can be formulated as
\begin{align}
(\boldsymbol{a}_l^h)^T\boldsymbol{W}\boldsymbol{a}_l^h
=&\mathop{\sum}\limits_{i\geq l'}w_{i}(x_{i}-\eta_{l})^2\nonumber\\
    =&\mathop{\sum}\limits_{i\geq l'}w_{i}x_i^2-2w_ix_i\eta_l+w_i\eta_l^2\nonumber\\
    =&\mathop{\sum}\limits_{i\geq l'}w_{i}x_i^2-2\eta_l\mathop{\sum}\limits_{i\geq l'}w_ix_i+\eta_l^2\mathop{\sum}\limits_{i\geq l'}w_i. 
\label{f2}
\end{align}
Following from (\ref{f2}), the value of $(\boldsymbol{a}_l^h)^T\boldsymbol{W}\boldsymbol{a}_l^h$ is determined by $\eta_l$ and the values of accumulating $w_ix_i^2$, $w_ix_i$ and $w_i$ from $l'$ to $N$.
We note that for different $l$'s, we can share the value of the accumulated $w_ix_i^2$, $w_ix_i$ and $w_i$ to reduce the computational complexity.
The pseudo-code of computing $(\boldsymbol{a}_l^h)^T\boldsymbol{W}\boldsymbol{a}_l^h$ for all $l$'s is summarized in  Algorithm~\ref{alg:norm}. 
Specifically, in the first step of Algorithm~\ref{alg:norm}, we find the change point for the hinge function with each threshold $\eta_l$.
We then accumulate the reversed $\{w_ix_i^2\}$, $w_ix_i$, and $w_i$ for all the possible values of $l'$. 
In our implementation, this operation is performed efficiently by invoking Numpy built-in function ``cumsum'' with linear computational complexity.
In the final step, we evaluate $(\boldsymbol{a}_l^h)^T\boldsymbol{W}\boldsymbol{a}_l^h$
according to (\ref{f2}).
The computation of $(\boldsymbol{a}^r_l)^T\boldsymbol{W}\boldsymbol{a}_l^r$ can be implemented similarly.

\RestyleAlgo{ruled}

%% This is needed if you want to add comments in
%% your algorithm with \Comment
\SetKwComment{Comment}{/* }{ */}
\begin{algorithm}
\caption{Compute weighted norm}\label{alg:norm}
\KwInput{Dataset $\{(x_i,r_i,w_i)\}_{i=1}^N$, threshold set $\{\eta_l\}_{l=1}^L$}
\KwOutput{$\{(\boldsymbol{a}_l^h)^T\boldsymbol{W}\boldsymbol{a}_l^h\}_{l=1}^L$}
Step 1: for each $l$, find $l'$ such that $x_{l'-1}<\eta_l\le x_{l'}$\\
Step 2: reverse $\{x_i\}$ and $\{w_i\}$\\
Step 3: $\boldsymbol{z}^o = \text{cumsum}(\{w_ix_i^2\})$, $\boldsymbol{z}^p = \text{cumsum}(\{w_ix_i\})$, and $\boldsymbol{z}^w = \text{cumsum}(\{w_i\})$\\
Step 4: for each $l$, $(\boldsymbol{a}_l^h)^T\boldsymbol{W}\boldsymbol{a}_l^h=z_{l'}^o-2\eta_lz^p_{l'}+\eta_l^2z^w_{l'}$\\ 
\end{algorithm}

\begin{figure*}[t]
	\centering
\subfigure[Sample weight modification interface.]{
    \includegraphics[width=8cm]{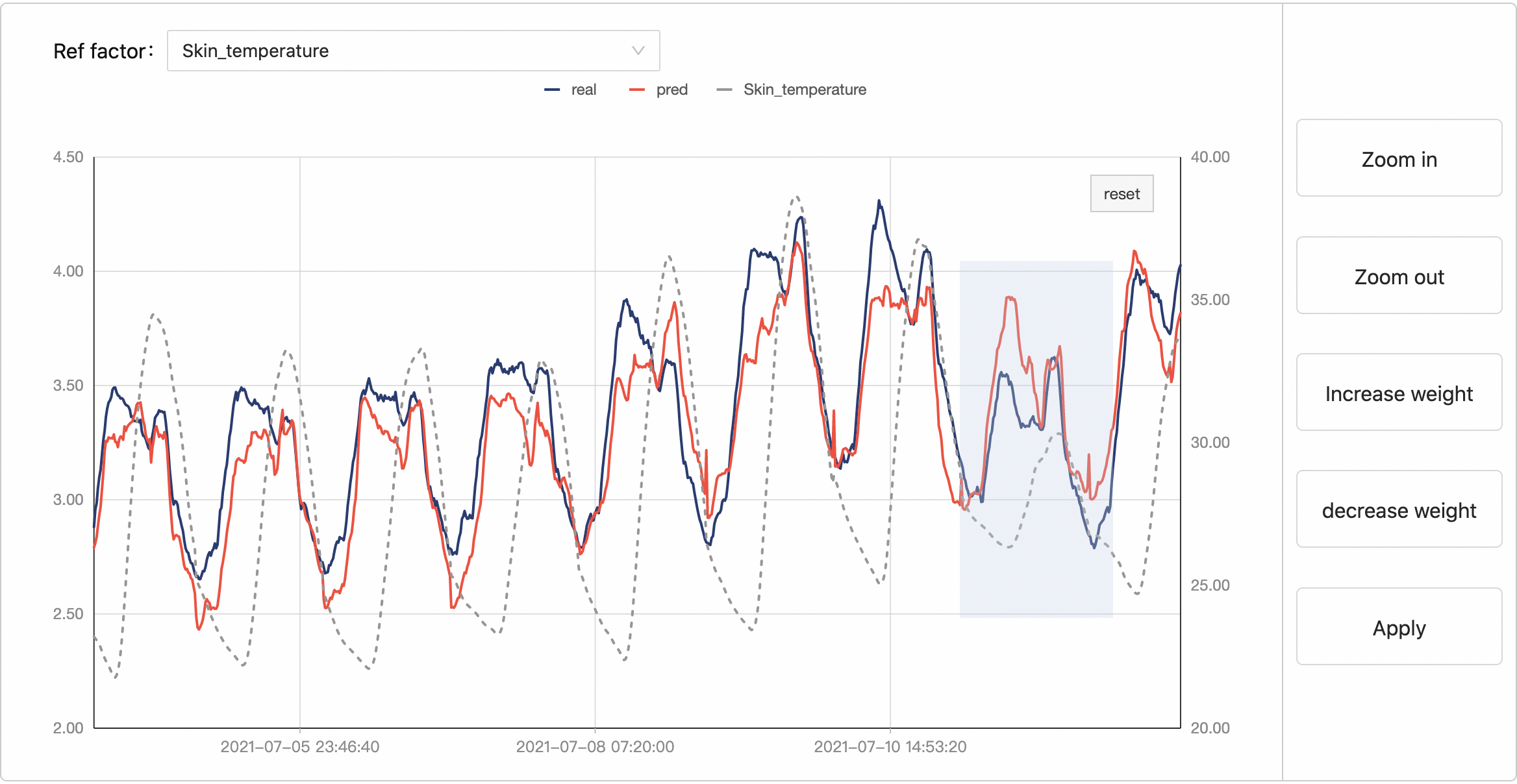}\label{fig:usr-interface-1}
}
\subfigure[Shape function modification interface.]{
    \includegraphics[width=8cm]{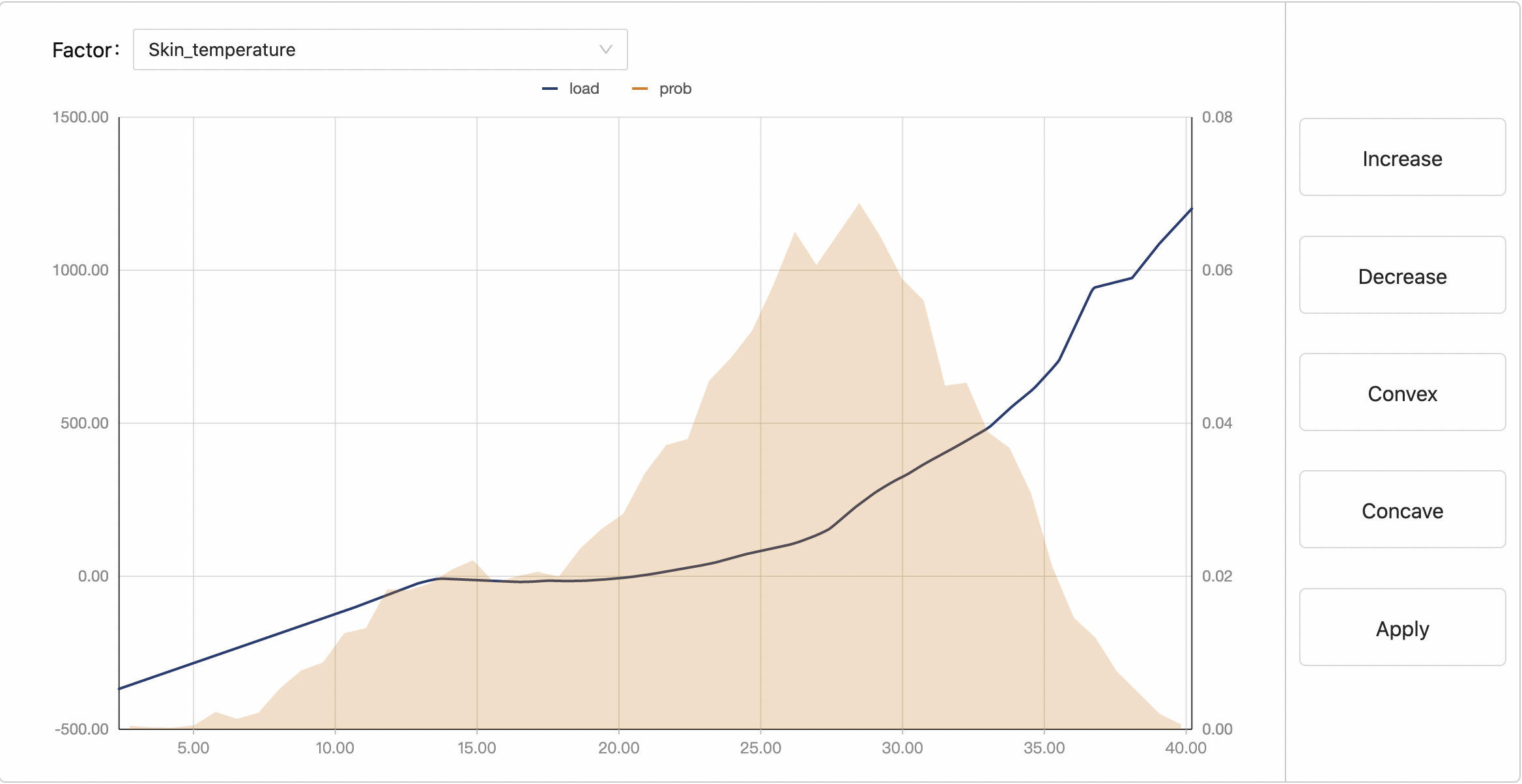}\label{fig:usr-interface-2}
}
\vspace{-0.3cm}
	\caption{User interface of our interactive GAM.}
\vspace{-0.3cm}	
\end{figure*}

Similar to $(\boldsymbol{a}^h_l)^T\boldsymbol{W}\boldsymbol{a}_l^h$, $\boldsymbol{b}^T\boldsymbol{W}\boldsymbol{a}_l^h$ can be formulatetd as
\begin{align}
\boldsymbol{b}^T\boldsymbol{W}\boldsymbol{a}_l^h
=&\mathop{\sum}\limits_{i\geq l'}w_{i}b_i(x_{i}-\eta_{l})\nonumber\\
    =&\mathop{\sum}\limits_{i\geq l'}w_{i}b_ix_i-w_{i}b_i\eta_l\nonumber\\
    =&\mathop{\sum}\limits_{i\geq l'}w_{i}b_ix_i-\eta_l\mathop{\sum}\limits_{i\geq l'}w_{i}b_i. 
\label{f3}
\end{align}
We can also compute the accumulation of $w_{i}b_ix_i$ and $w_{i}b_i$ in
advance to reduce the computational complexity.
The Pseudo-code of efficiently computing weighted correlation $\boldsymbol{b}^T\boldsymbol{W}\boldsymbol{a}_l^h$ is shown in Algorithm~\ref{alg:corr}. 

So far, we have shown how to compute $(\boldsymbol{a}^h_l)^T\boldsymbol{W}\boldsymbol{a}_l^h$ and 
$\boldsymbol{b}^T\boldsymbol{W}\boldsymbol{a}_l^h$ efficiently,
which enables us to implement Step 1 in Algorithm~\ref{alg:omp} in linear time 
complexity. Note that the proposed methods not only reduce the computational complexity, but also reduce the space complexity by avoiding explicitly generating matrix $\boldsymbol{A}\in\mathbb{R}^{N\times(2L+1)}$, which is space-consuming when $N$ and $L$ are large.

\RestyleAlgo{ruled}

%% This is needed if you want to add comments in
%% your algorithm with \Comment
\SetKwComment{Comment}{/* }{ */}

\begin{algorithm}
\caption{Compute weighted correlation}\label{alg:corr}
\KwInput{Dataset $\{(x_i,b_i,w_i)\}_{i=1}^N$, threshold set $\{\eta_l\}_{l=1}^L$}
\KwOutput{$\{\boldsymbol{b}^T\boldsymbol{W}\boldsymbol{a}_l^h\}_{l=1}^L$}
 Step 1: for each $l$, find $l'$ such that $x_{l'-1}<\eta_l\le x_{l'}$\\
Step 2: reverse $\{x_i\}$, $\{b_i\}$, and $\{w_i\}$\\
Step 3: $\boldsymbol{z}^o = \text{cumsum}(\{w_ix_ib_i\})$ and $\boldsymbol{z}^p = \text{cumsum}(\{w_ib_i\})$\\
Step 4: for each $l$, $\boldsymbol{b}^T\boldsymbol{W}\boldsymbol{a}_l^h=z_{l'}^o-\eta_lz^p_{l'}$\\ 
\end{algorithm}

% \begin{equation}
% \begin{aligned}

% \caption{}
% \label{eq:A}
%     r_i=\sum_{j\geq i}w_{j}(x_{j}-a_{i})^2\\
%     =\sum_{j\geq i}w_{j}(x_{j}-a_{i+1}+a_{i+1}-a_{i})^2\\
%     =\sum_{j\geq i}w_{j}(x_{j}-a_{i+1})^2+2w_{j}(x_{j}-a_{i+1})()
% \end{aligned}
% \end{equation}

% \subsubsection{Discussion}
% \vspace{-.7cm}
\subsection{User Interaction}
We first introduce how to impose the four types of constraints to represent the domain knowledge mathematically, and then introduce the web-based user interface which supports two types of operations, including data point importance adjustment and constraint imposing. 

% The expert users input two types of requests. 
% One tells the algorithm which part of the
% data points they concern. 
% The other restricts the trend of the load with respect to some factors.
% The former request are fulfilled by  upweighting the concerned data points.
% The latter one is translated to constraints. As introduced in Section \ref{sec:iii}, four types of constraints are considered in this paper, i.e., monotone increase, monotone decrease, convex and concave. By arbitrarily combining these four constraints to different extend, users can implement complex requests to the model.

%\subsubsection{Deal with constraints}
\subsubsection{Imposing Constraints to GAM}
Note that the feasible sets of these four types of constraints introduced in Section~\ref{sec:iii} are all convex. 
Recall the updating strategy of $f_d(x)$ shown in (\ref{eq:function-update}), we can treat the piecewise linear function $g(x)$ as the gradient of $f_d(x)$ and update $f_d(x)$ using gradient projection method to force $f_d(x)$ satisfying
the desired constraints, i.e.,
\begin{align}
    f_d^{(t)}(x) = \alpha f_d^{(t-1)}(x) + (1-\alpha)\mathcal{P}_d(f_d^{(t-1)}(x)+\mu g(x)),\label{eq:function-update-proj}
\end{align}
where the projection operator $\mathcal{P}_d(f)$ projects $f$ onto the feasible set of the constraints applied on the $i$th feature.

Without loss of generality, we next introduce how to project a function $f$ to the feasible set of monotone increasing constraints. 
Let $\boldsymbol{x}$ be a vector formed by the anchor points in the region of interest. 
We discuss how to find a monotone increasing vector $\boldsymbol{x}'$ such that $\|\boldsymbol{x}'-\boldsymbol{x}\|$ is minimized.
Formally, the vector $\boldsymbol{x}'$ can be found by solving the following problem
\begin{align}
    \min_{\boldsymbol{x}'}\; \|\boldsymbol{x}'-\boldsymbol{x}\|_2^2\qquad
    \text{s.t.}\; \boldsymbol{D}\boldsymbol{x}'\ge \boldsymbol{0},\label{problem:projection}
\end{align}
where the matrix $\boldsymbol{D}$ is the differential matrix.
Problem (\ref{problem:projection}) can be solved
by existing optimization methods, such as alternating direction multiplier methods (ADMM)~\cite{boyd:admm}.
Nevertheless, as the projection with quadratic complexity is invoked in every iteration, it is unpractical to repeat it. 
To speed up the computation, we devise an approximation algorithm.
Specifically, $\boldsymbol{x}'$ is set to be
the average of the monotone increasing upper envelope and the lower envelope.
If $\boldsymbol{x}$ is already monotone increasing, our method will ensure   $\boldsymbol{x}'=\boldsymbol{x}$, which is a key property of the projection operator.

The rest three constraints can be handled similarly. For monotone decreasing constraint, we reverse the vector, apply the monotone increasing approximation and then reverse it back. 
For convex constraints, we find a vector $\boldsymbol{x}'$, whose difference is monotone increasing. We approximate the difference of $\boldsymbol{x}$ using our monotone increasing approximate method, and then accumulate it back
to get the convex approximation.
Similarly, concave constraints can be satisfied by forcing the difference of $\boldsymbol{x}$ to be monotone decreasing.

\subsubsection{User interface}
For the convenience of experts with no experience in machine learning to incorporate domain knowledge into GAM, we develop a web-based user interface for our interactive GAM. 
The interface supports two types of operations.
% domain knowledge.

The first type of operation adjusts the importance of the data points. In ELF, users usually care more about prediction under extreme weather events (such as extremely low or high temperature). 
They can ask the model to pay more attention to historical events with specific extreme weather conditions. 
To this end, our interactive GAM provides a sample weight modification interface, as shown in Fig.\ref{fig:usr-interface-1} . 
The weight modification interface plots the historical and the predicted electric load in ``real'' and ``pred'' lines, respectively (the scale of the load on the left is distorted due to data privacy). 
To increase the weight, the user only needs to select a segment of the lines and click the ``Increase weight'' button
to upweight selected samples. 
Each button click will multiply the weights of the selected samples by $2$. 
We also provide ``Decrease weight'' button for users to downweights certain outliers.
After the weights are modified, users can click the button ``Apply'' to retrain the model. 
To help the user find similar events quickly, we also plot the value of the reference factors to indicate the type of the extreme event in dash line, labeled as the factor name, such as ``Skin\_temperature'', with scale on the right.
Users can select the reference factor from the drop down box named ``Ref Factor''. 
As the extreme events rarely occur, we also provide ``Zoom in'' and ``Zoom out'' button to help the user search similar events over a long time range.

The second type of operation adjusts the GAM model by imposing the four types of constraints discussed in Section~\ref{sec:iii}. 
% is the ``trend knowledge'' as discussed in section \ref{sec:iii}.
We develop the shape function modification interface, as shown in Figure~\ref{fig:usr-interface-2}. The line in the main view is the shape function of the specified factor labeled as ``load'', with its scale on the left. The shaded part labelled as ``prob'' is the data density that indicates how many data will be affected if constraints are added on this area, with its scale on the right. Users can first select a target region and choose an option from
``Increase'', ``Decrease'', ``Convex'' and ``Concave'' to impose monotone increasing, monotone decreasing, convex
or concave constraints onto the selected area, respectively.
The factor can be chosen from the drop down box named ``Factor''. 
Multiple constraints can be applied on one or different shape functions.
Once all the constraints are imposed, by clicking
``Apply'' button, both the model and the shape functions will be updated to satisfy the user-imposed constraints. 
Although constraints are only imposed on a fraction of shape functions, all the shape functions will be updated to minimize the fitting error.

\section{Experiments}

\subsection{Experiments on Public Benchmark Datasets}
In this subsection, we evaluate our proposed method in benchmark regression tasks. Table \ref{tab:uci} summarizes the selected 6 public datasets, including "Abalone" ~\cite{abalone}, "Ailerons"~\cite{aileron}, "Boston Housing"~\cite{boston}, "Pole"~\cite{pole}, "Stock"~\cite{stock} and  "ComputerAct"~\cite{computeract}.

\begin{table}[htbp]
\vspace{-.3cm}
\caption{Statistics of public benchmark datasets.}
\vspace{-.4cm}
\begin{center}
\begin{tabular}{c c c c}
\hline
\textbf{Dataset}&\textbf{Size}&\textbf{Attributes}&\textbf{Data Source}\\
\hline

Abalone&4177&8&UCI\\
Ailerons&13750&40&Rui Camacho\\
Boston Housing&506&13&UCI\\
Pole&15000&48&UCI\\
Stock&950&10&StatLib\\
ComputerAct&8192&22&DELVE\\

\hline
\end{tabular}
%\vspace{-5pt}
\label{tab:uci}
\end{center}
\end{table}

% \begin{table*}[htbp]
% \caption{MSE of Public Dataset}
% \begin{center}
% \begin{tabular}{c c c c c c c}
% \hline
% \textbf{Model}&\textbf{Abalone}&\textbf{Ailerons}&\textbf{Boston Housing}&\textbf{Pole}&\textbf{Stock}&\textbf{ComputerAct}\\
% \hline
% \textbf{Ours}&\textbf{0.4164}&\textbf{0.1744}&\textbf{0.1402}&\textbf{458.8043}&\textbf{0.0026}&\textbf{0.02179}\\
% EBM&0.4729&0.1837&0.1656&460.5753&0.0114&0.02186\\
% pyGAM&0.4272&0.1936&0.1582&464.1290&0.0142&0.02207\\

% \hline
% \end{tabular}
% %\vspace{-5pt}
% \label{tab:comp_acc}
% \end{center}
% \end{table*}

\begin{table*}[htbp]
\caption{MSE of different algorithms on public benchmark datasets.}
% \vspace{-.4cm}
\begin{center}
\begin{tabular}{c c c c c c c}
\hline
\textbf{Model}&\textbf{Abalone}&\textbf{Ailerons}&\textbf{Boston Housing}&\textbf{Pole}&\textbf{Stock}&\textbf{ComputerAct}\\
\hline
\textbf{Ours}&\textbf{0.4164}&\textbf{0.1744}&\textbf{0.1402}&\textbf{458.8043}&\textbf{0.0026}&\textbf{0.02179}\\
EBM&0.4729&0.1837&0.1656&460.5753&0.0114&0.02186\\
pyGAM&0.4272&0.1936&0.1582&464.1290&0.0142&0.02207\\
FLAM&0.4224&0.1822&0.2120&478.0053&0.0066&0.02206\\

\hline
\end{tabular}
%\vspace{-5pt}
\label{tab:comp_acc}
\end{center}
\end{table*}

% \begin{figure*}
%     \centering
% \subfigure[Interactive GAM]{
%     \includegraphics[width=0.31\linewidth]{Editable GAM.pdf}
% }
% \bigskip
% \centering
% \subfigure[EBM]{
% \begin{minipage}[t]{0.32\textwidth}
% \centering
% \includegraphics[width=1\linewidth]{EBM.pdf}
% \end{minipage}
% }
% \subfigure[pyGAM]{
% \begin{minipage}[t]{0.32\textwidth}
% \centering
% \includegraphics[width=1\linewidth]{GAM.pdf}
% \end{minipage}
% }
% \vspace{-.7cm}
% \caption{The predicting results on dataset ``Stock''}
% \label{fig:test_result}
% \end{figure*}

\begin{figure*}
    \centering
    \includegraphics[width=0.24\linewidth]{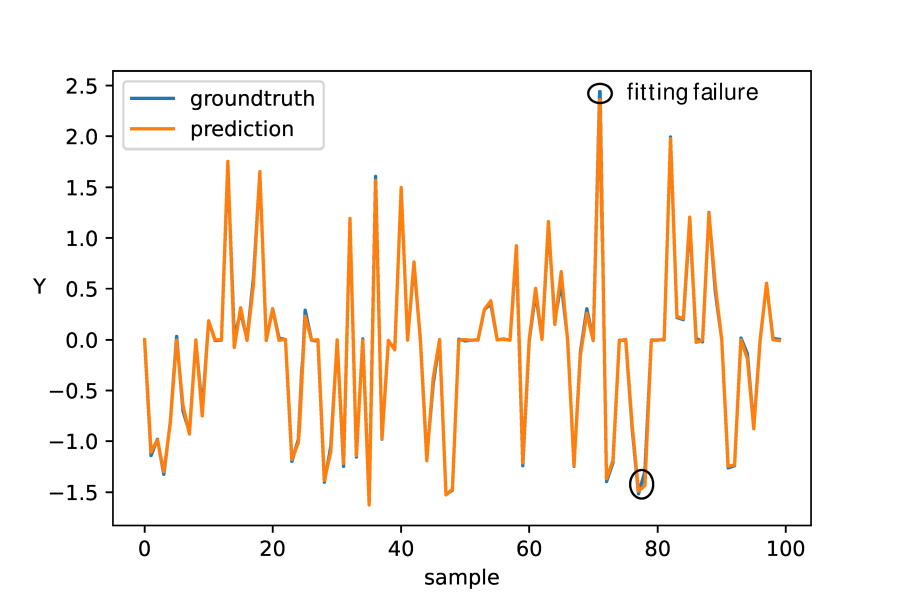}
\includegraphics[width=0.24\linewidth]{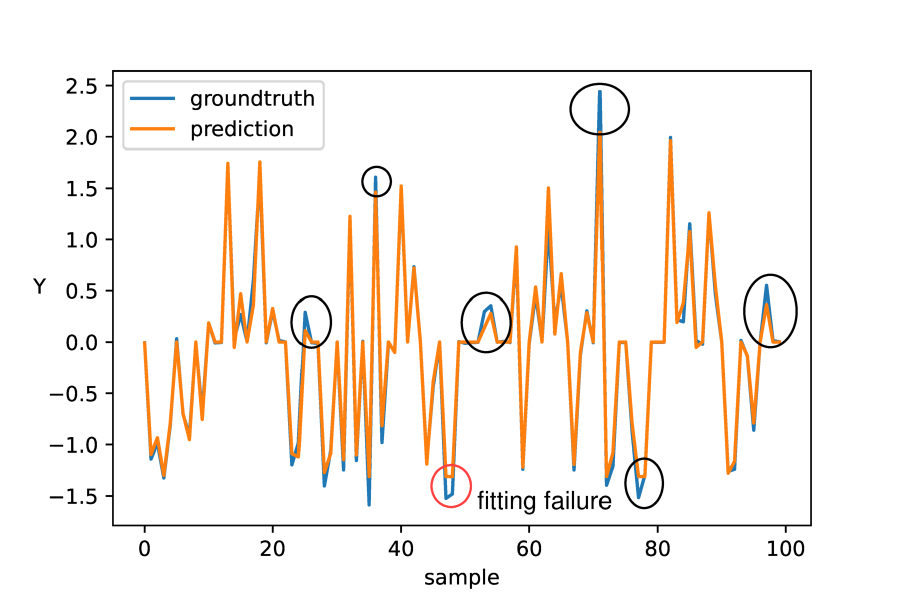}
\includegraphics[width=0.24\linewidth]{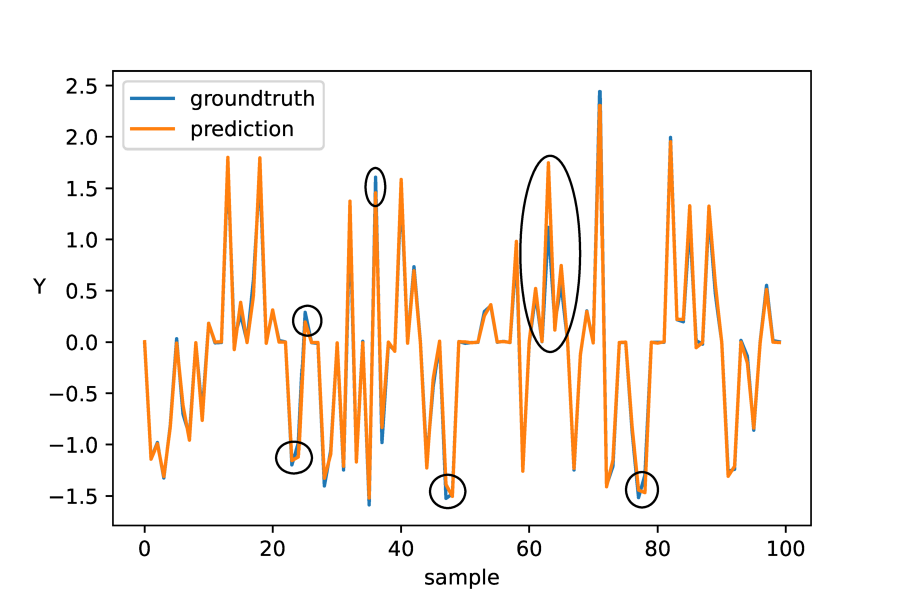}
\includegraphics[width=0.22\linewidth]{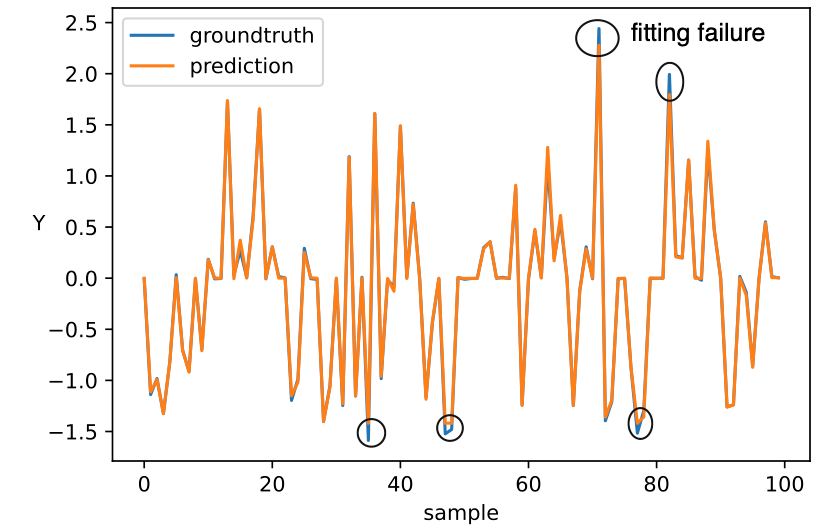}
\caption{The predicting results on Stock dataset. From left to right: prediction obtained by our piecewise linear GAM, EBM, pyGAM, FLAM, respectively.}
% \vspace{-0.4cm}
\label{fig:test_result}
\end{figure*}

% \begin{figure*}
%     \centering
% \subfigure[Our method]{
%     \includegraphics[width=0.32\linewidth]{E-GAMFeature1.png}
% \label{fig:GAM}
% }
% \bigskip
% \centering
% \subfigure[EBM]{
% \begin{minipage}[t]{0.32\textwidth}
% \centering
% \includegraphics[width=1\linewidth]{EBMfeature1.png}
% \end{minipage}
% \label{fig:ebm}
% }
% \subfigure[pyGAM]{
% \begin{minipage}[t]{0.32\textwidth}
% \centering
% \includegraphics[width=1\linewidth]{total_feature.pdf}
% \end{minipage}
% \label{fig:pyGAM}
% }
% \caption{The Response Function for three algorithms}
% \label{fig:reponse}
% \end{figure*}

\begin{figure}
    \centering
\includegraphics[width=1\linewidth]{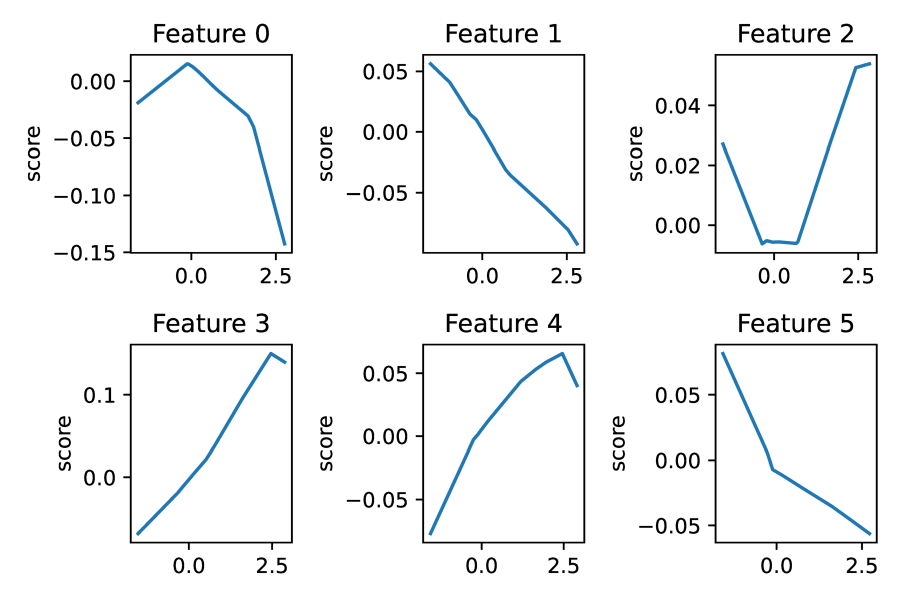}
% \vspace{-.6cm}
\caption{The shape functions of the proposed method learned on Stock dataset. }
\label{fig:reponse}
\end{figure}
% \vspace{-.3cm}

% \begin{figure*}
%     \centering
% \subfigure[Editable GAM]{
%     \includegraphics[width=0.32\linewidth]{GAM-group.pdf}
% \label{fig:GAM}
% }
% \bigskip
% \centering
% \subfigure[EBM]{
% \begin{minipage}[t]{0.32\textwidth}
% \centering
% \includegraphics[width=1\linewidth]{EBM-group.png}
% \end{minipage}
% \label{fig:ebm}
% }
% \subfigure[pyGAM]{
% \begin{minipage}[t]{0.32\textwidth}
% \centering
% \includegraphics[width=1\linewidth]{pyGAM-group.pdf}
% \end{minipage}
% \label{fig:pyGAM}
% }
% \caption{The Response Function for three algorithms}
% \label{fig:reponse}
% \end{figure*}

\subsubsection{Experiment Setup}
% All the algorithms are implemented in Python. We leverage the scikit learn package~\cite{sklearn} to calculate the least squares solution in our algorithm.
Three state-of-the-art GAM algorithms are compared, including EBM~\cite{nori2019interpretml}, pyGAM~\cite{serven2018pygam},
and FLAM~\cite{petersen2019data}.
EBM\footnote{Code available at \url{https://github.com/interpretml/interpret}} is a GAM based on boosted-tree, which
% We leverage the open-source python implementation of EBM, interpret-ml~\cite{nori2019interpretml}, to finish the experiment evaluation. 
assumes the shape functions of features
are piecewise-constant.
EBM learns the model by iteratively generating 
decision trees, where 
only one feature can be used for splitting
at each tree.
% The EBM begins by bucketing data from continuous features into discrete bins, ensuring that each bin has approximately equal amounts of data(piecewise constant). The procedure is a common optimization in tree-based learning algorithms, is also used by LightGBM and XGBoost~\cite{ke2017lightgbm,chen2016xgboost}. Then, EBM learns a decision tree is finding the best split for every feature. After finishing the pre-processing, EBM uses cyclic gradient boosting to learn shape functions for each feature. To enforce additivity, each boosting tree is only related to one feature. 
pyGAM\footnote{Code available at \url{https://github.com/dswah/pyGAM}} is a widely used python GAM package,
where shape functions are modeled as low-order splines. Backfitting method is used to learn the paramters of the splines.
FLAM\footnote{Code available at \url{https://cran.r-project.org/web/packages/flam/}} is a convex optimization based method,
where the shape function is represented as a piecewise constant function and Fused LASSO~\cite{tibshirani2005sparsity} is invoked to solve the problem.
Thoughtout our experiments, we set
$\lambda=1$, $K=7$, and $\mu=0.1$. We turn off the interaction and bagging
in EBM for fair comparison.

% pyGAM is a package for building Generalized Additive Models in Python. It is traditionally fit using smooth low-order splines. Mathematically speaking, GAM is a modeling technique where the impact of the predictive variables is captured through smooth functions. We could describe the pyGAM structure as \ref{eq:gam}: 
% \begin{align}
%     g(Y) = \beta + s_1(x_1)+...+s_d(x_d)
%     \label{eq:gam}
% \end{align}

% where $Y$ is the variable that are we are trying to predict, g(Y) denotes the link function that links the expected value to the feature $x_1,...,x_d$. The terms $s_1(x),...,s_d(x)$ denote smooth and low-order splines functions.
% FLAM is a flexible and interpretable additive regression model. It aims to avoid pre-specifying the functional form of the conditional association between each covariate and the response, while still retaining interpretability of the fitted functions.
% Thoughtout our experiments, we set
% $\lambda=1$, $K=7$, and $\mu=0.1$ for our
% method. We leverage LinearGAM in pyGAM package to fit our dataset. We turn off the interaction and bagging for fair comparsion. 
% All experiments are performed on a ThinkPad with Intel Core i7-10750H CPU @ 2.60GHz and 16GB RAM.

\subsubsection{Evaluation}
Mean square error (MSE) is used to evaluate performance:
\begin{align}
    \text{MSE} = \frac{1}{N}\sum_{i=1}^N(y_i-\hat{y}_i)^2, 
\end{align}
where $\hat{y}_i$ is the predicted value.
For all experiments, we leverage 5-fold cross-validation and calculate the average MSE for the evaluation result. From Table \ref{tab:comp_acc}, 
our algorithm outperforms other two algorithms in all 6 tasks, especially with enourmous improvement~(77\%) on the Stock dataset.

We take ``Stock'' as an example to analyze why our model outperforms other algorithms. Figure \ref{fig:test_result} shows the three algorithm's fitting results on the same testing dataset, with fitting gap marked by circles. Compared with Interactive GAM, EBM cannot well fit the peak and dip. We attribute the phenomenon to the limitation of piecewise constant. Unlike piecewise linear, piecewise constant will let a group of different feature values be the same result, which loses accuracy. Our inference can be verified through inspecting the marked red circles. EBM fits these areas with the same value, whereas in the groundtruth these values vary a lot. 
We also plot the shape functions of the first six features 
on dataset ``Stock'' learned by our method, as Figure \ref{fig:reponse} shows. Users can better understand the predicting mechanism of the model from the visualization of the shape functions.
% We show the EBM's response function in figure~\ref{fig:EBM}. To be intuitive, we also mark the red circle area's response function value. Due to the limitation of piecewise constant, the marked area of EBM's response function outputs the same result for every sample and leads to fitting failure. In the figure~\ref{fig:GAM}, our algorithm could perceive the feature's variety in the area and give more accurate fitting result. Compared with pyGAM, our response function is more smooth, which helps our model to fit the result better. We show a group of our algorithm's response functions in figure~\ref{fig:pyGAM}. 

\begin{table}[htbp]
\vspace{-.4cm}
\caption{RNMSE of different algorithms on the ELF dataset.}
\vspace{-.4cm}
\begin{center}
\begin{tabular}{c c c c c c}
\hline
\textbf{Model}&\textbf{Ours}&\textbf{EBM}&\textbf{PyGAM}&\textbf{FLAM}&\textbf{recency}\\
\hline
Overall &0.0859&0.0878&0.0893&0.0860&0.3645\\
Extreme event&0.0723&0.0736&0.0807&0.0937&0.0999\\
\hline
\end{tabular}
\vspace{-.4cm}
% \vspace{-5pt}
\label{tab:mape}
\end{center}
\end{table}

\begin{figure*}
    \centering
\subfigure[Fitting error of GAMs]{
    \includegraphics[width=0.18\linewidth]{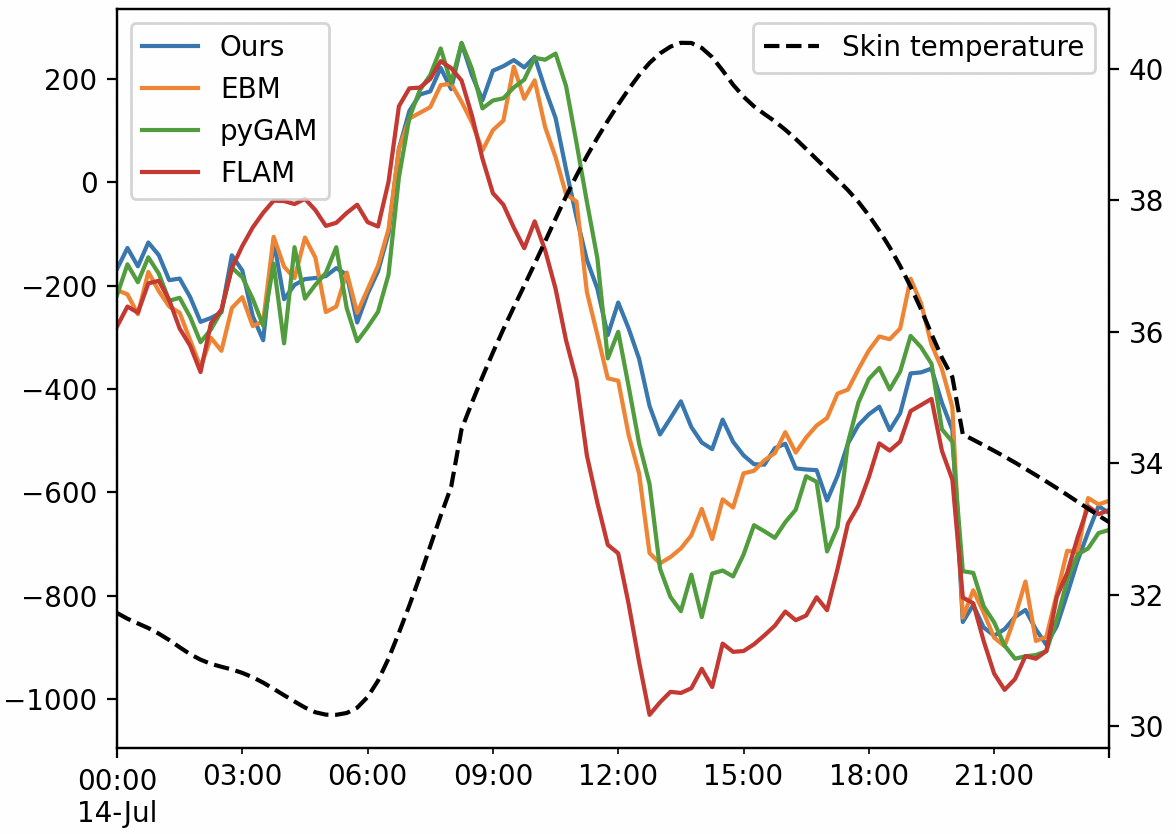}
}
\subfigure[Shape function learned by our method]{
\begin{minipage}[t]{0.18\textwidth}
\centering
\includegraphics[width=1\linewidth]{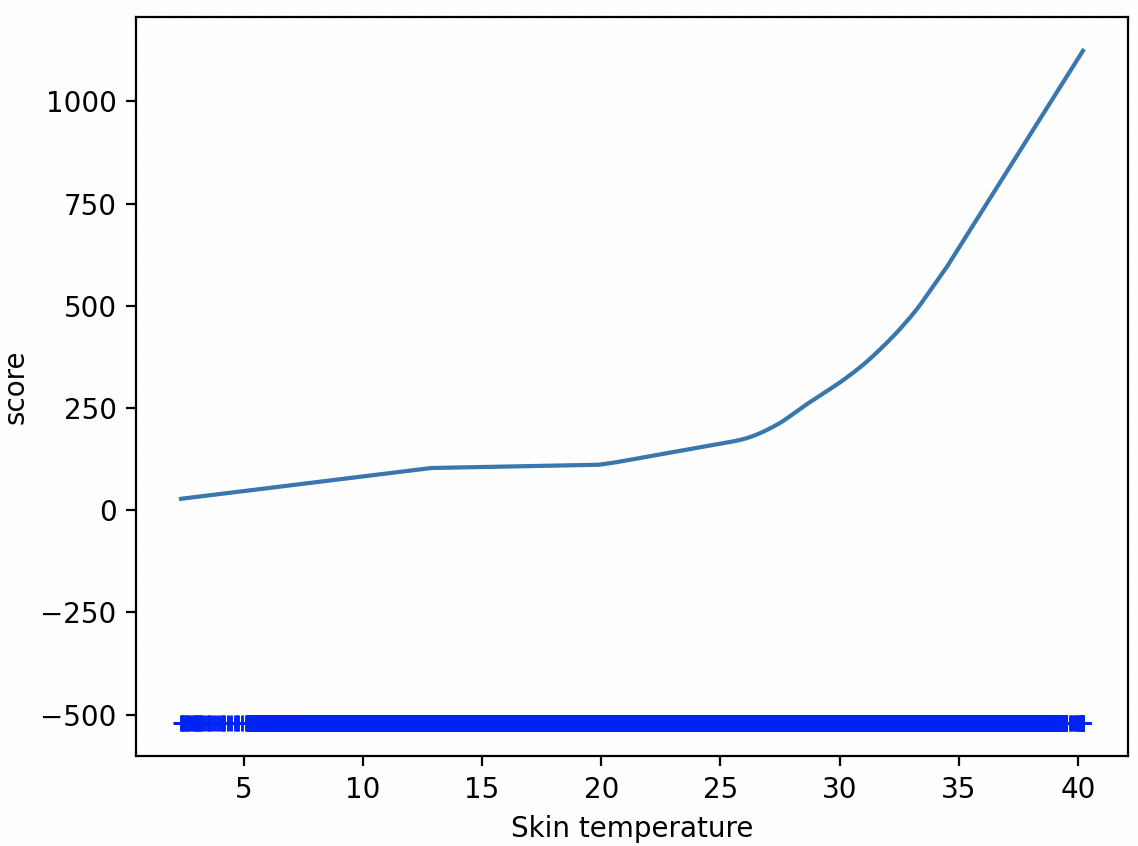}
\end{minipage}
}
\bigskip
\centering
\subfigure[Shape function learned by EBM]{
\begin{minipage}[t]{0.18\textwidth}
\centering
\includegraphics[width=1\linewidth]{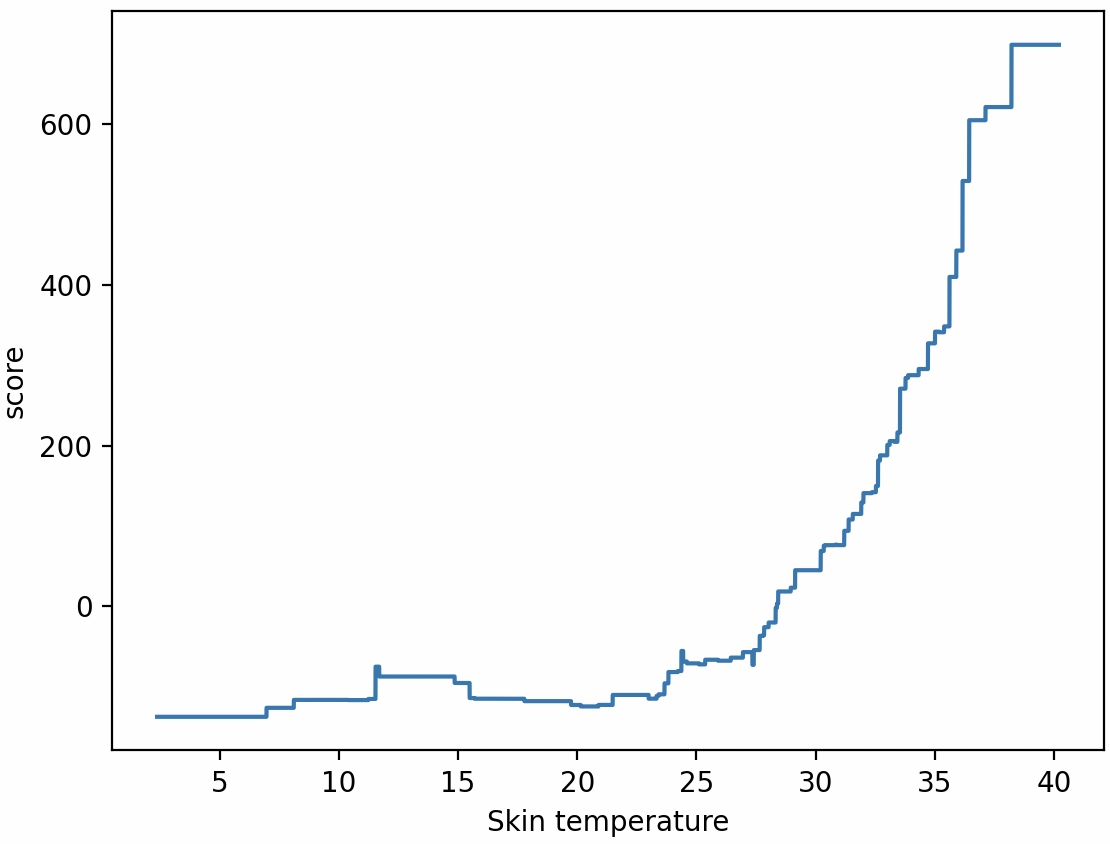}
\end{minipage}
}
\subfigure[Shape function learned by PyGAM]{
\begin{minipage}[t]{0.18\textwidth}
\centering
\includegraphics[width=1\linewidth]{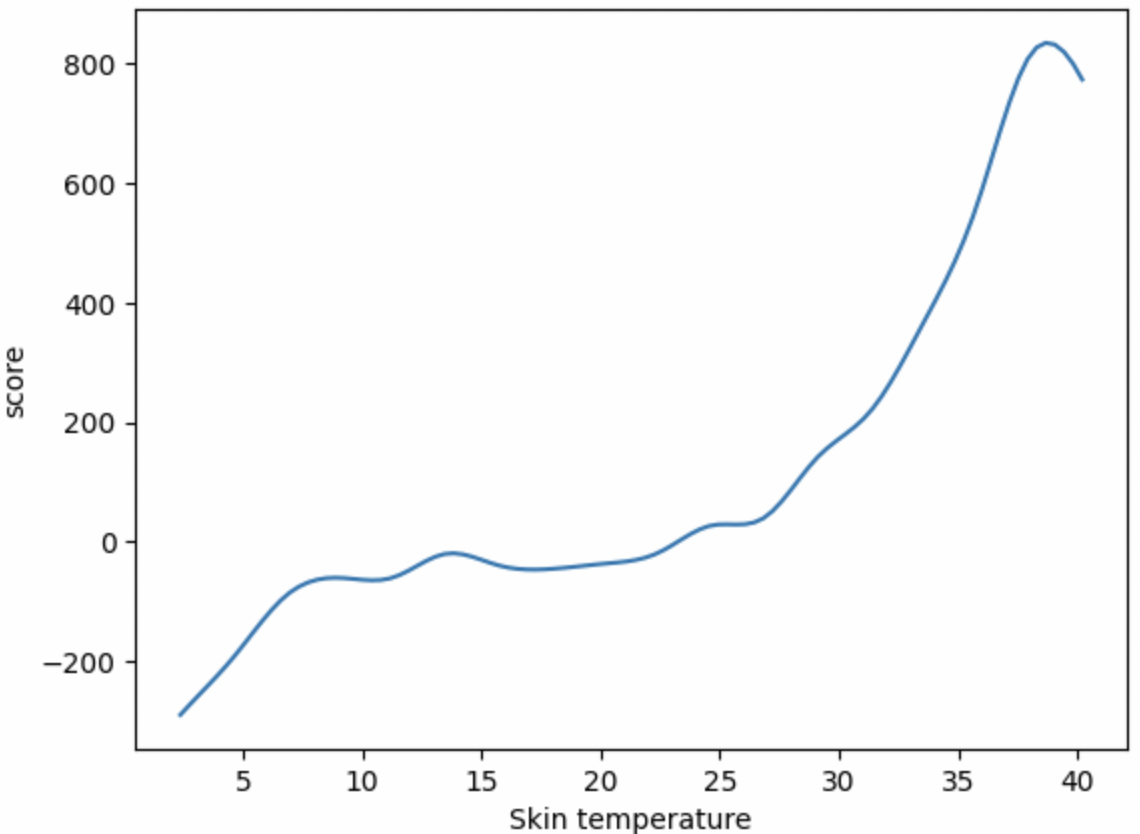}
\end{minipage}
}
\subfigure[Shape function learned by FLAM]{
\begin{minipage}[t]{0.18\textwidth}
\centering
\includegraphics[width=1\linewidth]{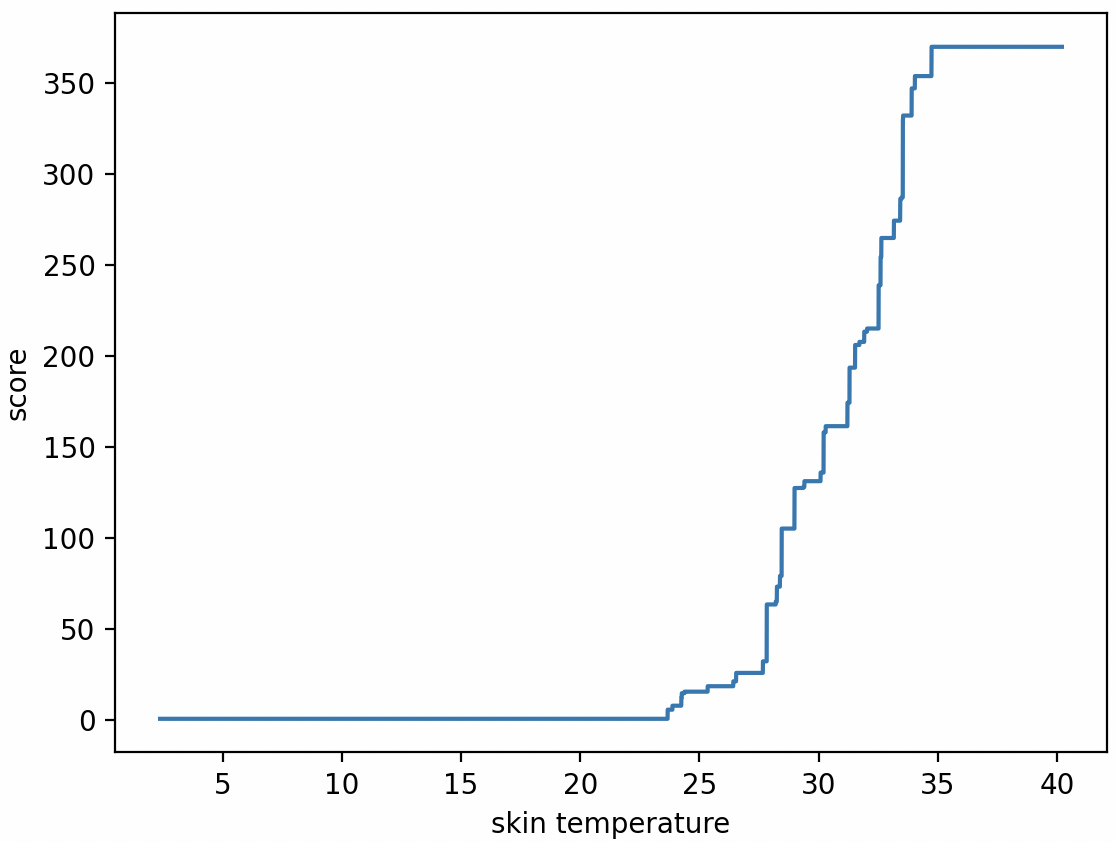}
\end{minipage}
}
 \vspace{-.7cm}
\caption{Fitting error and the shape functions of ``Skin tempturature'' learned by respective GAMs. }
\label{fig:fitting_error_ori}
\end{figure*}

\begin{figure*}
    \centering
\subfigure[Fitting error of GAMs]{
    \includegraphics[width=0.27\linewidth]{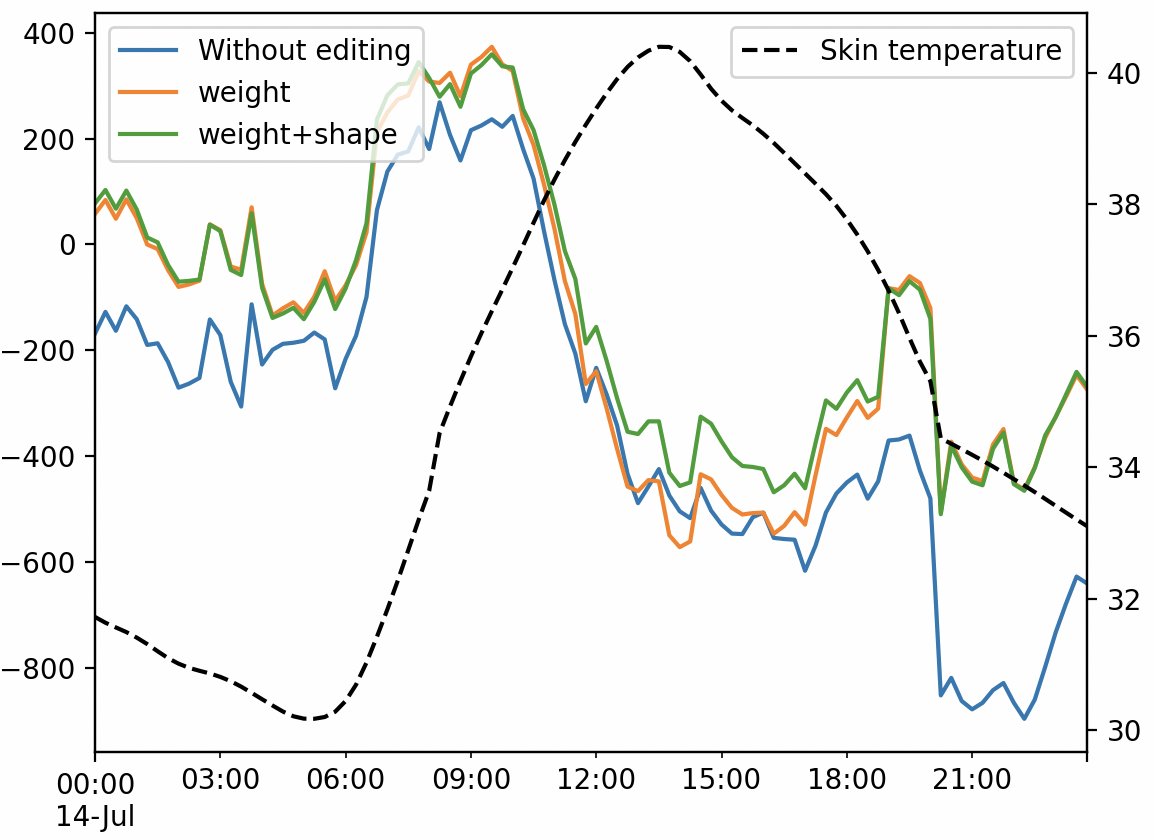}
}
\subfigure[Shape function after editing sample weights]{
\begin{minipage}[t]{0.27\textwidth}
\centering
\includegraphics[width=1\linewidth]{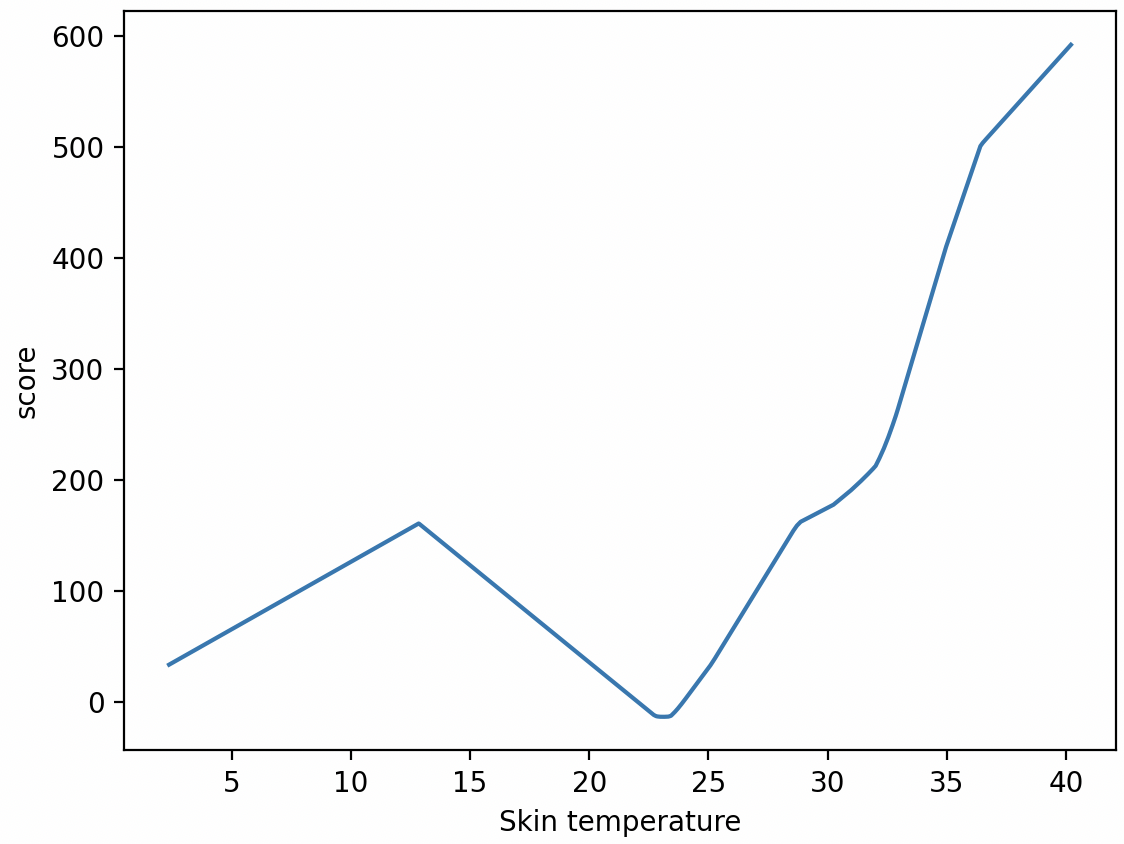}
\end{minipage}
}
\bigskip
\centering
\subfigure[Shape function after apply constraints]{
\begin{minipage}[t]{0.27\textwidth}
\centering
\includegraphics[width=1\linewidth]{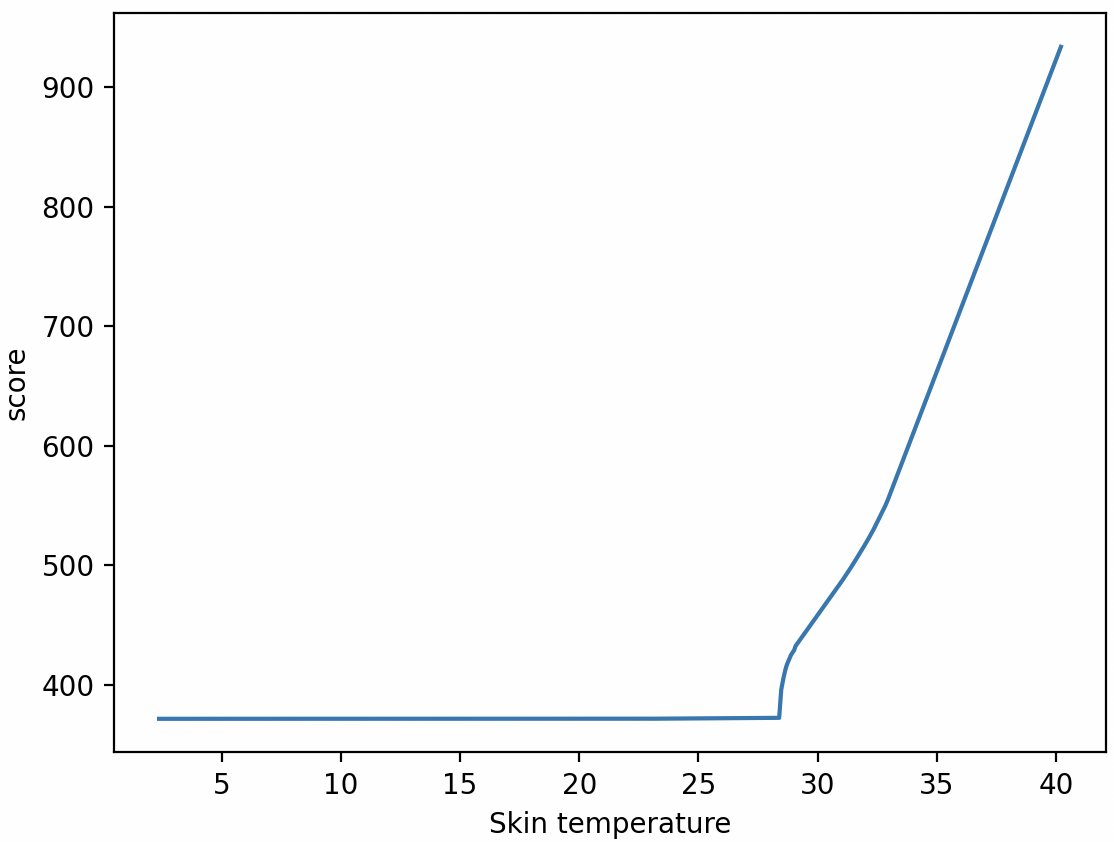}
\end{minipage}
}
\vspace{-.7cm}
\caption{Fitting error and 
the shape functions of ``Skin tempturature'' after editing our model.}
\label{fig:fitting_error_edit}
\end{figure*}
% \vspace{-.4cm}
\subsection{Experiments on Electric Load forecasting}
\label{sec:exp_elf}

In this subsection, we utilize a representative real-world electric load dataset to test the validity of our method. 
We also demonstrate how the experts can add domain knowledge to improve the performance of the model.
\subsubsection{Electric Load Dataset}
We collect the electric load data as well as the corresponding numerical weather prediction (NWP) of Henan province in China ranging
from February 11, 2020 to July 25, 2022.
We particularly focus on extreme weather events in summer, thus we only keep
the data from May 1 to October 31 each year.
The load data is recorded every 15 minutes, so there are 96
records a day. The sample size is 43488.
The provincial NWP data is obtained from city-level averages. 
We use 6 key NWP features in prediction, including surface pressure, surface sensible
heat flux, total cloud cover, surface net solar radiation,
total precipitation, and skin temperature\footnote{The detail meanings of these parameters can be found at \url{https://www.ecmwf.int/en/forecasts}}.
In reality, if skin temperature is high for days (even if unchanged), then
the electric load will increase gradually, known as ``accumulating effect''.
To model this effect, lagged features of skin temperature are generated.
Apart from weather information, we also generate calendar- and solar-based features to capture the periodic property of the load. They are time-of-the-day,
day-of-the-week, and month-of-the-year at the recorded time, respectively capturing the daily,
weekly, and monthly cycles. We also generate features to indicate holiday effect. 
Lagged features of load are also generated as baseline reference.
There are 18 features in total in the ELF dataset.
We use the data before July 1, 2022 
as training set and the rest as test set.
We note that in July 14, 2022, Henan province suffers from
extremely high temperature weather where 
the highest average skin temperature is above $40^{\circ}$C. We use the data on that day to measure the generalization ability under extreme weather events of each method.

\subsubsection{Performance Evaluation}
In addition to GAMs we compared above,
we also consider
a state-of-the-art ELF method, marked as ``recency''~\cite{wang2016electric}, which models ``recency effect'' at an aggregated level. The ``recency'' method generates several features with power values of the
temperature, and assumes that the electric load is linear
in these features. Unlike GAMs that contain no interactions, the ``recency'' method includes a few of second order interactions.
% We note that we don't include the results
% of FLAM as we cannot complete the experiments
% in one hour.
We set $\lambda=0.1$, $K=5$, $\mu=0.05$ and $\alpha=0.1$ for our method,
and use default settings for the rest methods.
We measure the performance of respective methods using root mean normalized square error (RNMSE):
\begin{align}
    \text{RNMSE} = \sqrt{\frac{1}{N}\sum_{i=1}^N\frac{(y_i-\hat{y}_i)^2}{y_i^2}},
\end{align}
where $\hat{y}_i$ denotes the predicted value.

In Table \ref{tab:mape}, our proposed method outperforms the rest in terms of RNMSE. To measure the generalization ability on extreme events, we also report the RNMSE on July 14, 2022, when extreme high temperature occured, as Table \ref{tab:mape} shows. Our proposed method reaches the lowest RNMSE when making this extrapolation.
Figure \ref{fig:fitting_error_ori}(a) reports the fitting error, namely the gap between the predicted load and the true load on that day. All the methods have similar fitting error, except from 12:00
to 16:00, when the skin temperature is extremely high. Our proposed method produces fitting errors relatively closer to zero. Figure \ref{fig:fitting_error_ori}(b)-\ref{fig:fitting_error_ori}(e) are the shape functions
learned by GAMs.
 Although all the GAMs are able to discovery the trend that higher temperature leads to
higher load, when the temperature is higher than $39^{\circ}$C, EBM, PyGAM 
and FLAM all fail to capture the increasing trend, which explains why our proposed method outperforms
the rest during the time from 12:00 to 16:00.

% Thus the testing error
% over the data after July 1, 2022 is able to measure the 
% generalization ability of the methods on extreme weather event.

\begin{table}[htbp]
% \vspace{-.3cm}
\caption{RNMSE of the proposed method before and after editing on extreme high temperature. }
\vspace{-.4cm}
\begin{center}
\begin{tabular}{c c c}
\hline
\textbf{No editing}&\textbf{Edit weight}&\textbf{Edit weight $+$ shape function}\\
\hline
0.0723 &0.0520&0.0475\\
\hline
\end{tabular}
%\vspace{-5pt}
% \vspace{-.6cm}
\label{tab:mape_edit}
\end{center}
\end{table}

\subsubsection{Evaluation of Interactive Functions} To evaluate how performance is improved through our built-in interactive functions, we edit our model on extreme weather events using our user
interface. We first select days with similar extreme events and perform the ``increase weight'' operation 16 times to retrain the model. In Figure \ref{fig:fitting_error_edit}(a), the orange line labeled as ``weight''  shows the fitting error after upweighting. From the figure we see that the fitting error
before 3:00 and
after 20:00 are significantly reduced. This is because the model pays more attention
to the extreme events in history and is able to capture the the pattern of the load under
these extreme events. Table \ref{tab:mape_edit} reports that after upweighting, the RNMSE is reduced from 0.0723 to 0.0520. From Figure \ref{fig:fitting_error_edit}(b), it is worth noticing that as we put
higher weight on the data with extremely high temperature, the shape function will fail to match the trend at low temperature. This issue can be fixed by restricting the curve from $0^{\circ}$C to $25^{\circ}$C to be monotone increasing. We apply the monotone increasing constraint and the re-updated model is further improved, as shown in Figure \ref{fig:fitting_error_edit}(a) with the green line labeled as ``shape function + weight''. Table \ref{tab:mape_edit} implies that with more domain knowledge involved, the RNMSE is further reduced from $0.0520$ to $0.0475$, which is about
$54\%$ of the RNMSE of EBM. Meanwhile, the shape function successfully agrees with the experts' domain knowledge in Figure \ref{fig:fitting_error_edit}.

\vspace{-0.15cm}
\section{Conclusion}
\vspace{-0.15cm}
In this paper, we propose an interactive GAM for electric load forecasting, allowing human experts to integrate domain knowledge in the model. While machine learning methods commonly overlook minority data such as extreme weather events and make unreliable extrapolations at out-of-the-box level, our method guarantees good performance within the range of factor values that users care about. We design an efficient algorithm to learn our interactive GAM composed of piecewise linear functions. The numerical results on public benchmark data and ELF data show that our method not only gives outperforming accuracy, but also extrapolates well at extreme factor levels and exhibits strong generalization ability. The method is proved to be practical in China, as users can easily refine the model with their expert knowledge by using the web-based interactive interface we deployed. 

\vspace{-0.2cm}
\begin{acks}
This work was supported by Alibaba Group through Alibaba Research Intern Program.
\end{acks}

\bibliographystyle{ACM-Reference-Format}
\bibliography{acmart}

%%% -*-BibTeX-*-
%%% Do NOT edit. File created by BibTeX with style
%%% ACM-Reference-Format-Journals [18-Jan-2012].

\begin{thebibliography}{60}

%%% ====================================================================
%%% NOTE TO THE USER: you can override these defaults by providing
%%% customized versions of any of these macros before the \bibliography
%%% command.  Each of them MUST provide its own final punctuation,
%%% except for \shownote{}, \showDOI{}, and \showURL{}.  The latter two
%%% do not use final punctuation, in order to avoid confusing it with
%%% the Web address.
%%%
%%% To suppress output of a particular field, define its macro to expand
%%% to an empty string, or better, \unskip, like this:
%%%
%%% \newcommand{\showDOI}[1]{\unskip}   % LaTeX syntax
%%%
%%% \def \showDOI #1{\unskip}           % plain TeX syntax
%%%
%%% ====================================================================

\ifx \showCODEN    \undefined \def \showCODEN     #1{\unskip}     \fi
\ifx \showDOI      \undefined \def \showDOI       #1{#1}\fi
\ifx \showISBNx    \undefined \def \showISBNx     #1{\unskip}     \fi
\ifx \showISBNxiii \undefined \def \showISBNxiii  #1{\unskip}     \fi
\ifx \showISSN     \undefined \def \showISSN      #1{\unskip}     \fi
\ifx \showLCCN     \undefined \def \showLCCN      #1{\unskip}     \fi
\ifx \shownote     \undefined \def \shownote      #1{#1}          \fi
\ifx \showarticletitle \undefined \def \showarticletitle #1{#1}   \fi
\ifx \showURL      \undefined \def \showURL       {\relax}        \fi
% The following commands are used for tagged output and should be
% invisible to TeX
\providecommand\bibfield[2]{#2}
\providecommand\bibinfo[2]{#2}
\providecommand\natexlab[1]{#1}
\providecommand\showeprint[2][]{arXiv:#2}

\bibitem[ELF(2021)]%
        {ELFbook:2021}
 \bibinfo{year}{2021}\natexlab{}.
\newblock In \bibinfo{booktitle}{\emph{Intelligent Data-Analytics for Condition Monitoring}}, \bibfield{editor}{\bibinfo{person}{Hasmat Malik}, \bibinfo{person}{Nuzhat Fatema}, {and} \bibinfo{person}{Atif Iqbal}} (Eds.). \bibinfo{publisher}{Academic Press}, \bibinfo{pages}{iv}.
\newblock
\showISBNx{978-0-323-85510-5}
\urldef\tempurl%
\url{https://doi.org/10.1016/B978-0-323-85510-5.00012-0}
\showDOI{\tempurl}


\bibitem[{Abalone}(1995)]%
        {abalone}
\bibfield{author}{\bibinfo{person}{{Abalone}}.} \bibinfo{year}{1995}\natexlab{}.
\newblock \bibinfo{howpublished}{Rui Camacho}.
\newblock
\urldef\tempurl%
\url{https://archive-beta.ics.uci.edu/dataset/1/abalone}
\showURL{%
\tempurl}


\bibitem[Abbasi et~al\mbox{.}(2019)]%
        {abbasi2019short}
\bibfield{author}{\bibinfo{person}{Raza~Abid Abbasi}, \bibinfo{person}{Nadeem Javaid}, \bibinfo{person}{Muhammad Nauman~Javid Ghuman}, \bibinfo{person}{Zahoor~Ali Khan}, {and} \bibinfo{person}{Shujat Ur~Rehman}.} \bibinfo{year}{2019}\natexlab{}.
\newblock \showarticletitle{Short term load forecasting using XGBoost}. In \bibinfo{booktitle}{\emph{Web, Artificial Intelligence and Network Applications: Proceedings of the Workshops of the 33rd International Conference on Advanced Information Networking and Applications (WAINA-2019) 33}}. Springer, \bibinfo{pages}{1120--1131}.
\newblock


\bibitem[{Aileron}(1995)]%
        {aileron}
\bibfield{author}{\bibinfo{person}{{Aileron}}.} \bibinfo{year}{1995}\natexlab{}.
\newblock \bibinfo{howpublished}{UCI Machine Learning Repository}.
\newblock
\urldef\tempurl%
\url{https://archive-beta.ics.uci.edu/dataset/1/aileron}
\showURL{%
\tempurl}


\bibitem[Bauer and Kohavi(1999)]%
        {bauer1999empirical}
\bibfield{author}{\bibinfo{person}{Eric Bauer} {and} \bibinfo{person}{Ron Kohavi}.} \bibinfo{year}{1999}\natexlab{}.
\newblock \showarticletitle{An empirical comparison of voting classification algorithms: Bagging, boosting, and variants}.
\newblock \bibinfo{journal}{\emph{Machine learning}}  \bibinfo{volume}{36} (\bibinfo{year}{1999}), \bibinfo{pages}{105--139}.
\newblock


\bibitem[{Boston Housing}(1988)]%
        {stock}
\bibfield{author}{\bibinfo{person}{{Boston Housing}}.} \bibinfo{year}{1988}\natexlab{}.
\newblock \bibinfo{howpublished}{StatLib}.
\newblock
\urldef\tempurl%
\url{https://www.dcc.fc.up.pt/~ltorgo/Regression/stock.html}
\showURL{%
\tempurl}


\bibitem[{Boston Housing}(1995)]%
        {boston}
\bibfield{author}{\bibinfo{person}{{Boston Housing}}.} \bibinfo{year}{1995}\natexlab{}.
\newblock \bibinfo{howpublished}{Delve Datasets}.
\newblock
\urldef\tempurl%
\url{https://www.cs.toronto.edu/~delve/data/boston/}
\showURL{%
\tempurl}


\bibitem[Boyd et~al\mbox{.}(2011)]%
        {boyd:admm}
\bibfield{author}{\bibinfo{person}{Stephen Boyd}, \bibinfo{person}{Neal Parikh}, \bibinfo{person}{Eric Chu}, \bibinfo{person}{Borja Peleato}, {and} \bibinfo{person}{Jonathan Eckstein}.} \bibinfo{year}{2011}\natexlab{}.
\newblock \showarticletitle{Distributed Optimization and Statistical Learning via the Alternating Direction Method of Multipliers}.
\newblock \bibinfo{journal}{\emph{Found. Trends Mach. Learn.}} \bibinfo{volume}{3}, \bibinfo{number}{1} (\bibinfo{date}{jan} \bibinfo{year}{2011}), \bibinfo{pages}{1–122}.
\newblock
\showISSN{1935-8237}


\bibitem[Chen and Guestrin(2016)]%
        {chen2016xgboost}
\bibfield{author}{\bibinfo{person}{Tianqi Chen} {and} \bibinfo{person}{Carlos Guestrin}.} \bibinfo{year}{2016}\natexlab{}.
\newblock \showarticletitle{Xgboost: A scalable tree boosting system}. In \bibinfo{booktitle}{\emph{Proceedings of the 22nd ACM SIGKDD international conference on knowledge discovery and data mining}}. \bibinfo{pages}{785--794}.
\newblock


\bibitem[Chen et~al\mbox{.}(2022)]%
        {chen2022learning}
\bibfield{author}{\bibinfo{person}{Weiqi Chen}, \bibinfo{person}{Wenwei Wang}, \bibinfo{person}{Bingqing Peng}, \bibinfo{person}{Qingsong Wen}, \bibinfo{person}{Tian Zhou}, {and} \bibinfo{person}{Liang Sun}.} \bibinfo{year}{2022}\natexlab{}.
\newblock \showarticletitle{Learning to Rotate: Quaternion Transformer for Complicated Periodical Time Series Forecasting}. In \bibinfo{booktitle}{\emph{Proceedings of the 28th ACM SIGKDD Conference on Knowledge Discovery and Data Mining}}. \bibinfo{pages}{146--156}.
\newblock


\bibitem[{ComputerActivity}(1995)]%
        {computeract}
\bibfield{author}{\bibinfo{person}{{ComputerActivity}}.} \bibinfo{year}{1995}\natexlab{}.
\newblock \bibinfo{howpublished}{Delve Datasets}.
\newblock
\urldef\tempurl%
\url{https://www.openml.org/search?type=data&sort=runs&id=197&status=active}
\showURL{%
\tempurl}


\bibitem[Dryar(1944)]%
        {dryar1944effect}
\bibfield{author}{\bibinfo{person}{Henry~A Dryar}.} \bibinfo{year}{1944}\natexlab{}.
\newblock \showarticletitle{The effect of weather on the system load}.
\newblock \bibinfo{journal}{\emph{Electrical Engineering}} \bibinfo{volume}{63}, \bibinfo{number}{12} (\bibinfo{year}{1944}), \bibinfo{pages}{1006--1013}.
\newblock


\bibitem[D’Ambrosio et~al\mbox{.}(2010)]%
        {d2010piecewise}
\bibfield{author}{\bibinfo{person}{Claudia D’Ambrosio}, \bibinfo{person}{Andrea Lodi}, {and} \bibinfo{person}{Silvano Martello}.} \bibinfo{year}{2010}\natexlab{}.
\newblock \showarticletitle{Piecewise linear approximation of functions of two variables in MILP models}.
\newblock \bibinfo{journal}{\emph{Operations Research Letters}} \bibinfo{volume}{38}, \bibinfo{number}{1} (\bibinfo{year}{2010}), \bibinfo{pages}{39--46}.
\newblock


\bibitem[Fan and Hyndman(2011)]%
        {fan2011short}
\bibfield{author}{\bibinfo{person}{Shu Fan} {and} \bibinfo{person}{Rob~J Hyndman}.} \bibinfo{year}{2011}\natexlab{}.
\newblock \showarticletitle{Short-term load forecasting based on a semi-parametric additive model}.
\newblock \bibinfo{journal}{\emph{IEEE transactions on power systems}} \bibinfo{volume}{27}, \bibinfo{number}{1} (\bibinfo{year}{2011}), \bibinfo{pages}{134--141}.
\newblock


\bibitem[Fan and Hyndman(2012)]%
        {fan2012forecasting}
\bibfield{author}{\bibinfo{person}{Shu Fan} {and} \bibinfo{person}{Rob~J Hyndman}.} \bibinfo{year}{2012}\natexlab{}.
\newblock \showarticletitle{Forecasting electricity demand in australian national electricity market}. In \bibinfo{booktitle}{\emph{2012 IEEE Power and Energy Society General Meeting}}. IEEE, \bibinfo{pages}{1--4}.
\newblock


\bibitem[Friedman(2001)]%
        {friedman2001greedy}
\bibfield{author}{\bibinfo{person}{Jerome~H Friedman}.} \bibinfo{year}{2001}\natexlab{}.
\newblock \showarticletitle{Greedy function approximation: a gradient boosting machine}.
\newblock \bibinfo{journal}{\emph{Annals of statistics}} (\bibinfo{year}{2001}), \bibinfo{pages}{1189--1232}.
\newblock


\bibitem[Gaillard et~al\mbox{.}(2016)]%
        {gaillard2016additive}
\bibfield{author}{\bibinfo{person}{Pierre Gaillard}, \bibinfo{person}{Yannig Goude}, {and} \bibinfo{person}{Rapha{\"e}l Nedellec}.} \bibinfo{year}{2016}\natexlab{}.
\newblock \showarticletitle{Additive models and robust aggregation for GEFCom2014 probabilistic electric load and electricity price forecasting}.
\newblock \bibinfo{journal}{\emph{International Journal of forecasting}} \bibinfo{volume}{32}, \bibinfo{number}{3} (\bibinfo{year}{2016}), \bibinfo{pages}{1038--1050}.
\newblock


\bibitem[Goodfellow et~al\mbox{.}(2016)]%
        {GoodBengCour16}
\bibfield{author}{\bibinfo{person}{Ian~J. Goodfellow}, \bibinfo{person}{Yoshua Bengio}, {and} \bibinfo{person}{Aaron Courville}.} \bibinfo{year}{2016}\natexlab{}.
\newblock \bibinfo{booktitle}{\emph{Deep Learning}}.
\newblock \bibinfo{publisher}{MIT Press}, \bibinfo{address}{Cambridge, MA, USA}.
\newblock
\newblock
\shownote{\url{http://www.deeplearningbook.org}}.


\bibitem[Hastie et~al\mbox{.}(2009)]%
        {hastie2009elements}
\bibfield{author}{\bibinfo{person}{Trevor Hastie}, \bibinfo{person}{Robert Tibshirani}, \bibinfo{person}{Jerome~H Friedman}, {and} \bibinfo{person}{Jerome~H Friedman}.} \bibinfo{year}{2009}\natexlab{}.
\newblock \bibinfo{booktitle}{\emph{The elements of statistical learning: data mining, inference, and prediction}}. Vol.~\bibinfo{volume}{2}.
\newblock \bibinfo{publisher}{Springer}.
\newblock


\bibitem[Hastie(2017)]%
        {hastie2017generalized}
\bibfield{author}{\bibinfo{person}{Trevor~J Hastie}.} \bibinfo{year}{2017}\natexlab{}.
\newblock \showarticletitle{Generalized additive models}.
\newblock In \bibinfo{booktitle}{\emph{Statistical models in S}}. \bibinfo{publisher}{Routledge}, \bibinfo{pages}{249--307}.
\newblock


\bibitem[He(2017)]%
        {he2017load}
\bibfield{author}{\bibinfo{person}{Wan He}.} \bibinfo{year}{2017}\natexlab{}.
\newblock \showarticletitle{Load forecasting via deep neural networks}.
\newblock \bibinfo{journal}{\emph{Procedia Computer Science}}  \bibinfo{volume}{122} (\bibinfo{year}{2017}), \bibinfo{pages}{308--314}.
\newblock


\bibitem[Hegselmann et~al\mbox{.}(2020)]%
        {hegselmann2020evaluation}
\bibfield{author}{\bibinfo{person}{Stefan Hegselmann}, \bibinfo{person}{Thomas Volkert}, \bibinfo{person}{Hendrik Ohlenburg}, \bibinfo{person}{Antje Gottschalk}, \bibinfo{person}{Martin Dugas}, {and} \bibinfo{person}{Christian Ertmer}.} \bibinfo{year}{2020}\natexlab{}.
\newblock \showarticletitle{An evaluation of the doctor-interpretability of generalized additive models with interactions}. In \bibinfo{booktitle}{\emph{Machine Learning for Healthcare Conference}}. PMLR, \bibinfo{pages}{46--79}.
\newblock


\bibitem[Hong and Fan(2016)]%
        {hong2016probabilistic}
\bibfield{author}{\bibinfo{person}{Tao Hong} {and} \bibinfo{person}{Shu Fan}.} \bibinfo{year}{2016}\natexlab{}.
\newblock \showarticletitle{Probabilistic electric load forecasting: A tutorial review}.
\newblock \bibinfo{journal}{\emph{International Journal of Forecasting}} \bibinfo{volume}{32}, \bibinfo{number}{3} (\bibinfo{year}{2016}), \bibinfo{pages}{914--938}.
\newblock


\bibitem[Hong et~al\mbox{.}(2014)]%
        {hong2014global}
\bibfield{author}{\bibinfo{person}{Tao Hong}, \bibinfo{person}{Pierre Pinson}, {and} \bibinfo{person}{Shu Fan}.} \bibinfo{year}{2014}\natexlab{}.
\newblock \bibinfo{title}{Global energy forecasting competition 2012}.
\newblock , \bibinfo{numpages}{357--363}~pages.
\newblock


\bibitem[Hong et~al\mbox{.}(2019)]%
        {hong2019global}
\bibfield{author}{\bibinfo{person}{Tao Hong}, \bibinfo{person}{Jingrui Xie}, {and} \bibinfo{person}{Jonathan Black}.} \bibinfo{year}{2019}\natexlab{}.
\newblock \showarticletitle{Global energy forecasting competition 2017: Hierarchical probabilistic load forecasting}.
\newblock \bibinfo{journal}{\emph{International Journal of Forecasting}} \bibinfo{volume}{35}, \bibinfo{number}{4} (\bibinfo{year}{2019}), \bibinfo{pages}{1389--1399}.
\newblock


\bibitem[Huang and Perry(2016)]%
        {huang2016semi}
\bibfield{author}{\bibinfo{person}{Jing Huang} {and} \bibinfo{person}{Matthew Perry}.} \bibinfo{year}{2016}\natexlab{}.
\newblock \showarticletitle{A semi-empirical approach using gradient boosting and k-nearest neighbors regression for GEFCom2014 probabilistic solar power forecasting}.
\newblock \bibinfo{journal}{\emph{International Journal of Forecasting}} \bibinfo{volume}{32}, \bibinfo{number}{3} (\bibinfo{year}{2016}), \bibinfo{pages}{1081--1086}.
\newblock


\bibitem[Ke et~al\mbox{.}(2017)]%
        {ke2017lightgbm}
\bibfield{author}{\bibinfo{person}{Guolin Ke}, \bibinfo{person}{Qi Meng}, \bibinfo{person}{Thomas Finley}, \bibinfo{person}{Taifeng Wang}, \bibinfo{person}{Wei Chen}, \bibinfo{person}{Weidong Ma}, \bibinfo{person}{Qiwei Ye}, {and} \bibinfo{person}{Tie-Yan Liu}.} \bibinfo{year}{2017}\natexlab{}.
\newblock \showarticletitle{LightGBM: A Highly Efficient Gradient Boosting Decision Tree}. In \bibinfo{booktitle}{\emph{Advances in Neural Information Processing Systems}}, \bibfield{editor}{\bibinfo{person}{I.~Guyon}, \bibinfo{person}{U.~Von Luxburg}, \bibinfo{person}{S.~Bengio}, \bibinfo{person}{H.~Wallach}, \bibinfo{person}{R.~Fergus}, \bibinfo{person}{S.~Vishwanathan}, {and} \bibinfo{person}{R.~Garnett}} (Eds.), Vol.~\bibinfo{volume}{30}. \bibinfo{publisher}{Curran Associates, Inc.}
\newblock
\urldef\tempurl%
\url{https://proceedings.neurips.cc/paper/2017/file/6449f44a102fde848669bdd9eb6b76fa-Paper.pdf}
\showURL{%
\tempurl}


\bibitem[Khotanzad et~al\mbox{.}(1998)]%
        {khotanzad1998annstlf}
\bibfield{author}{\bibinfo{person}{Alireza Khotanzad}, \bibinfo{person}{Reza Afkhami-Rohani}, {and} \bibinfo{person}{Dominic Maratukulam}.} \bibinfo{year}{1998}\natexlab{}.
\newblock \showarticletitle{ANNSTLF-artificial neural network short-term load forecaster-generation three}.
\newblock \bibinfo{journal}{\emph{IEEE Transactions on Power Systems}} \bibinfo{volume}{13}, \bibinfo{number}{4} (\bibinfo{year}{1998}), \bibinfo{pages}{1413--1422}.
\newblock


\bibitem[Kuster et~al\mbox{.}(2017)]%
        {ELF:2017:survey}
\bibfield{author}{\bibinfo{person}{Corentin Kuster}, \bibinfo{person}{Yacine Rezgui}, {and} \bibinfo{person}{Monjur Mourshed}.} \bibinfo{year}{2017}\natexlab{}.
\newblock \showarticletitle{Electrical load forecasting models: A critical systematic review}.
\newblock \bibinfo{journal}{\emph{Sustainable Cities and Society}}  \bibinfo{volume}{35} (\bibinfo{year}{2017}), \bibinfo{pages}{257--270}.
\newblock
\showISSN{2210-6707}
\urldef\tempurl%
\url{https://doi.org/10.1016/j.scs.2017.08.009}
\showDOI{\tempurl}


\bibitem[Landry et~al\mbox{.}(2016)]%
        {landry2016probabilistic}
\bibfield{author}{\bibinfo{person}{Mark Landry}, \bibinfo{person}{Thomas~P Erlinger}, \bibinfo{person}{David Patschke}, {and} \bibinfo{person}{Craig Varrichio}.} \bibinfo{year}{2016}\natexlab{}.
\newblock \showarticletitle{Probabilistic gradient boosting machines for GEFCom2014 wind forecasting}.
\newblock \bibinfo{journal}{\emph{International Journal of Forecasting}} \bibinfo{volume}{32}, \bibinfo{number}{3} (\bibinfo{year}{2016}), \bibinfo{pages}{1061--1066}.
\newblock


\bibitem[Li et~al\mbox{.}(2017)]%
        {li2017building}
\bibfield{author}{\bibinfo{person}{Chengdong Li}, \bibinfo{person}{Zixiang Ding}, \bibinfo{person}{Dongbin Zhao}, \bibinfo{person}{Jianqiang Yi}, {and} \bibinfo{person}{Guiqing Zhang}.} \bibinfo{year}{2017}\natexlab{}.
\newblock \showarticletitle{Building energy consumption prediction: An extreme deep learning approach}.
\newblock \bibinfo{journal}{\emph{Energies}} \bibinfo{volume}{10}, \bibinfo{number}{10} (\bibinfo{year}{2017}), \bibinfo{pages}{1525}.
\newblock


\bibitem[Lou et~al\mbox{.}(2012)]%
        {lou2012intelligible}
\bibfield{author}{\bibinfo{person}{Yin Lou}, \bibinfo{person}{Rich Caruana}, {and} \bibinfo{person}{Johannes Gehrke}.} \bibinfo{year}{2012}\natexlab{}.
\newblock \showarticletitle{Intelligible models for classification and regression}. In \bibinfo{booktitle}{\emph{Proceedings of the 18th ACM SIGKDD international conference on Knowledge discovery and data mining}}. \bibinfo{pages}{150--158}.
\newblock


\bibitem[Lou et~al\mbox{.}(2013)]%
        {lou2013accurate}
\bibfield{author}{\bibinfo{person}{Yin Lou}, \bibinfo{person}{Rich Caruana}, \bibinfo{person}{Johannes Gehrke}, {and} \bibinfo{person}{Giles Hooker}.} \bibinfo{year}{2013}\natexlab{}.
\newblock \showarticletitle{Accurate intelligible models with pairwise interactions}. In \bibinfo{booktitle}{\emph{Proceedings of the 19th ACM SIGKDD international conference on Knowledge discovery and data mining}}. \bibinfo{pages}{623--631}.
\newblock


\bibitem[Magnani and Boyd(2009)]%
        {magnani2009convex}
\bibfield{author}{\bibinfo{person}{Alessandro Magnani} {and} \bibinfo{person}{Stephen~P Boyd}.} \bibinfo{year}{2009}\natexlab{}.
\newblock \showarticletitle{Convex piecewise-linear fitting}.
\newblock \bibinfo{journal}{\emph{Optimization and Engineering}}  \bibinfo{volume}{10} (\bibinfo{year}{2009}), \bibinfo{pages}{1--17}.
\newblock


\bibitem[Meyer and Pebesma(2021)]%
        {meyer2021predicting}
\bibfield{author}{\bibinfo{person}{Hanna Meyer} {and} \bibinfo{person}{Edzer Pebesma}.} \bibinfo{year}{2021}\natexlab{}.
\newblock \showarticletitle{Predicting into unknown space? Estimating the area of applicability of spatial prediction models}.
\newblock \bibinfo{journal}{\emph{Methods in Ecology and Evolution}} \bibinfo{volume}{12}, \bibinfo{number}{9} (\bibinfo{year}{2021}), \bibinfo{pages}{1620--1633}.
\newblock


\bibitem[Nedellec et~al\mbox{.}(2014)]%
        {nedellec2014gefcom2012}
\bibfield{author}{\bibinfo{person}{Raphael Nedellec}, \bibinfo{person}{Jairo Cugliari}, {and} \bibinfo{person}{Yannig Goude}.} \bibinfo{year}{2014}\natexlab{}.
\newblock \showarticletitle{GEFCom2012: Electric load forecasting and backcasting with semi-parametric models}.
\newblock \bibinfo{journal}{\emph{International Journal of forecasting}} \bibinfo{volume}{30}, \bibinfo{number}{2} (\bibinfo{year}{2014}), \bibinfo{pages}{375--381}.
\newblock


\bibitem[Nori et~al\mbox{.}(2019a)]%
        {ebm:2019}
\bibfield{author}{\bibinfo{person}{Harsha Nori}, \bibinfo{person}{Samuel Jenkins}, \bibinfo{person}{Paul Koch}, {and} \bibinfo{person}{Rich Caruana}.} \bibinfo{year}{2019}\natexlab{a}.
\newblock \bibinfo{title}{InterpretML: A Unified Framework for Machine Learning Interpretability}.
\newblock \bibinfo{howpublished}{ArXiv}.
\newblock
\urldef\tempurl%
\url{https://www.microsoft.com/en-us/research/publication/interpretml-a-unified-framework-for-machine-learning-interpretability/}
\showURL{%
\tempurl}


\bibitem[Nori et~al\mbox{.}(2019b)]%
        {nori2019interpretml}
\bibfield{author}{\bibinfo{person}{Harsha Nori}, \bibinfo{person}{Samuel Jenkins}, \bibinfo{person}{Paul Koch}, {and} \bibinfo{person}{Rich Caruana}.} \bibinfo{year}{2019}\natexlab{b}.
\newblock \showarticletitle{InterpretML: A Unified Framework for Machine Learning Interpretability}.
\newblock \bibinfo{journal}{\emph{arXiv preprint arXiv:1909.09223}} (\bibinfo{year}{2019}).
\newblock


\bibitem[Pati et~al\mbox{.}(1993)]%
        {omp}
\bibfield{author}{\bibinfo{person}{Yagyensh~Chandra Pati}, \bibinfo{person}{Ramin Rezaiifar}, {and} \bibinfo{person}{Perinkulam~Sambamurthy Krishnaprasad}.} \bibinfo{year}{1993}\natexlab{}.
\newblock \showarticletitle{Orthogonal matching pursuit: Recursive function approximation with applications to wavelet decomposition}. In \bibinfo{booktitle}{\emph{Proceedings of 27th Asilomar conference on signals, systems and computers}}. IEEE, \bibinfo{pages}{40--44}.
\newblock


\bibitem[Petersen and Witten(2019)]%
        {petersen2019data}
\bibfield{author}{\bibinfo{person}{Ashley Petersen} {and} \bibinfo{person}{Daniela Witten}.} \bibinfo{year}{2019}\natexlab{}.
\newblock \showarticletitle{Data-adaptive additive modeling}.
\newblock \bibinfo{journal}{\emph{Statistics in medicine}} \bibinfo{volume}{38}, \bibinfo{number}{4} (\bibinfo{year}{2019}), \bibinfo{pages}{583--600}.
\newblock


\bibitem[Petersen et~al\mbox{.}(2016)]%
        {petersen2016fused}
\bibfield{author}{\bibinfo{person}{Ashley Petersen}, \bibinfo{person}{Daniela Witten}, {and} \bibinfo{person}{Noah Simon}.} \bibinfo{year}{2016}\natexlab{}.
\newblock \showarticletitle{Fused lasso additive model}.
\newblock \bibinfo{journal}{\emph{Journal of Computational and Graphical Statistics}} \bibinfo{volume}{25}, \bibinfo{number}{4} (\bibinfo{year}{2016}), \bibinfo{pages}{1005--1025}.
\newblock


\bibitem[{Pole telecommunication}(1995)]%
        {pole}
\bibfield{author}{\bibinfo{person}{{Pole telecommunication}}.} \bibinfo{year}{1995}\natexlab{}.
\newblock \bibinfo{howpublished}{UCI Machine Learning Repository}.
\newblock
\urldef\tempurl%
\url{https://www.dcc.fc.up.pt/~ltorgo/Regression/pole.html}
\showURL{%
\tempurl}


\bibitem[Prokhorenkova et~al\mbox{.}(2018)]%
        {prokhorenkova2018catboost}
\bibfield{author}{\bibinfo{person}{Liudmila Prokhorenkova}, \bibinfo{person}{Gleb Gusev}, \bibinfo{person}{Aleksandr Vorobev}, \bibinfo{person}{Anna~Veronika Dorogush}, {and} \bibinfo{person}{Andrey Gulin}.} \bibinfo{year}{2018}\natexlab{}.
\newblock \showarticletitle{CatBoost: unbiased boosting with categorical features}.
\newblock \bibinfo{journal}{\emph{Advances in neural information processing systems}}  \bibinfo{volume}{31} (\bibinfo{year}{2018}).
\newblock


\bibitem[Rudin et~al\mbox{.}(2022)]%
        {RudinEtAlSurvey2022}
\bibfield{author}{\bibinfo{person}{Cynthia Rudin}, \bibinfo{person}{Chaofan Chen}, \bibinfo{person}{Zhi Chen}, \bibinfo{person}{Haiyang Huang}, \bibinfo{person}{Lesia Semenova}, {and} \bibinfo{person}{Chudi Zhong}.} \bibinfo{year}{2022}\natexlab{}.
\newblock \showarticletitle{{Interpretable machine learning: Fundamental principles and 10 grand challenges}}.
\newblock \bibinfo{journal}{\emph{Statistics Surveys}} \bibinfo{volume}{16}, \bibinfo{number}{none} (\bibinfo{year}{2022}), \bibinfo{pages}{1 -- 85}.
\newblock
\urldef\tempurl%
\url{https://doi.org/10.1214/21-SS133}
\showDOI{\tempurl}


\bibitem[Serv{\'e}n and Brummitt(2018)]%
        {serven2018pygam}
\bibfield{author}{\bibinfo{person}{Daniel Serv{\'e}n} {and} \bibinfo{person}{Charlie Brummitt}.} \bibinfo{year}{2018}\natexlab{}.
\newblock \showarticletitle{pygam: Generalized additive models in python}.
\newblock \bibinfo{journal}{\emph{Zenodo. doi}}  \bibinfo{volume}{10} (\bibinfo{year}{2018}).
\newblock


\bibitem[Shi et~al\mbox{.}(2017)]%
        {shi2017deep}
\bibfield{author}{\bibinfo{person}{Heng Shi}, \bibinfo{person}{Minghao Xu}, {and} \bibinfo{person}{Ran Li}.} \bibinfo{year}{2017}\natexlab{}.
\newblock \showarticletitle{Deep learning for household load forecasting—A novel pooling deep RNN}.
\newblock \bibinfo{journal}{\emph{IEEE Transactions on Smart Grid}} \bibinfo{volume}{9}, \bibinfo{number}{5} (\bibinfo{year}{2017}), \bibinfo{pages}{5271--5280}.
\newblock


\bibitem[Siahkamari et~al\mbox{.}(2020)]%
        {siahkamari2020piecewise}
\bibfield{author}{\bibinfo{person}{Ali Siahkamari}, \bibinfo{person}{Aditya Gangrade}, \bibinfo{person}{Brian Kulis}, {and} \bibinfo{person}{Venkatesh Saligrama}.} \bibinfo{year}{2020}\natexlab{}.
\newblock \showarticletitle{Piecewise linear regression via a difference of convex functions}. In \bibinfo{booktitle}{\emph{International Conference on Machine Learning}}. PMLR, \bibinfo{pages}{8895--8904}.
\newblock


\bibitem[Sigauke(2017)]%
        {sigauke2017forecasting}
\bibfield{author}{\bibinfo{person}{Caston Sigauke}.} \bibinfo{year}{2017}\natexlab{}.
\newblock \showarticletitle{Forecasting medium-term electricity demand in a South African electric power supply system}.
\newblock \bibinfo{journal}{\emph{Journal of Energy in Southern Africa}} \bibinfo{volume}{28}, \bibinfo{number}{4} (\bibinfo{year}{2017}), \bibinfo{pages}{54--67}.
\newblock


\bibitem[Smyl and Hua(2019)]%
        {smyl2019machine}
\bibfield{author}{\bibinfo{person}{Slawek Smyl} {and} \bibinfo{person}{N~Grace Hua}.} \bibinfo{year}{2019}\natexlab{}.
\newblock \showarticletitle{Machine learning methods for GEFCom2017 probabilistic load forecasting}.
\newblock \bibinfo{journal}{\emph{International Journal of Forecasting}} \bibinfo{volume}{35}, \bibinfo{number}{4} (\bibinfo{year}{2019}), \bibinfo{pages}{1424--1431}.
\newblock


\bibitem[Tan et~al\mbox{.}(2018)]%
        {tan2018distill}
\bibfield{author}{\bibinfo{person}{Sarah Tan}, \bibinfo{person}{Rich Caruana}, \bibinfo{person}{Giles Hooker}, {and} \bibinfo{person}{Yin Lou}.} \bibinfo{year}{2018}\natexlab{}.
\newblock \showarticletitle{Distill-and-compare: Auditing black-box models using transparent model distillation}. In \bibinfo{booktitle}{\emph{Proceedings of the 2018 AAAI/ACM Conference on AI, Ethics, and Society}}. \bibinfo{pages}{303--310}.
\newblock


\bibitem[Tibshirani et~al\mbox{.}(2005)]%
        {tibshirani2005sparsity}
\bibfield{author}{\bibinfo{person}{Robert Tibshirani}, \bibinfo{person}{Michael Saunders}, \bibinfo{person}{Saharon Rosset}, \bibinfo{person}{Ji Zhu}, {and} \bibinfo{person}{Keith Knight}.} \bibinfo{year}{2005}\natexlab{}.
\newblock \showarticletitle{Sparsity and smoothness via the fused lasso}.
\newblock \bibinfo{journal}{\emph{Journal of the Royal Statistical Society: Series B (Statistical Methodology)}} \bibinfo{volume}{67}, \bibinfo{number}{1} (\bibinfo{year}{2005}), \bibinfo{pages}{91--108}.
\newblock


\bibitem[Wang et~al\mbox{.}(2016)]%
        {wang2016electric}
\bibfield{author}{\bibinfo{person}{Pu Wang}, \bibinfo{person}{Bidong Liu}, {and} \bibinfo{person}{Tao Hong}.} \bibinfo{year}{2016}\natexlab{}.
\newblock \showarticletitle{Electric load forecasting with recency effect: A big data approach}.
\newblock \bibinfo{journal}{\emph{International Journal of Forecasting}} \bibinfo{volume}{32}, \bibinfo{number}{3} (\bibinfo{year}{2016}), \bibinfo{pages}{585--597}.
\newblock


\bibitem[Wang et~al\mbox{.}(2022)]%
        {wang2022interpretability}
\bibfield{author}{\bibinfo{person}{Zijie~J Wang}, \bibinfo{person}{Alex Kale}, \bibinfo{person}{Harsha Nori}, \bibinfo{person}{Peter Stella}, \bibinfo{person}{Mark~E Nunnally}, \bibinfo{person}{Duen~Horng Chau}, \bibinfo{person}{Mihaela Vorvoreanu}, \bibinfo{person}{Jennifer Wortman~Vaughan}, {and} \bibinfo{person}{Rich Caruana}.} \bibinfo{year}{2022}\natexlab{}.
\newblock \showarticletitle{Interpretability, Then What? Editing Machine Learning Models to Reflect Human Knowledge and Values}. In \bibinfo{booktitle}{\emph{Proceedings of the 28th ACM SIGKDD Conference on Knowledge Discovery and Data Mining}}. \bibinfo{pages}{4132--4142}.
\newblock


\bibitem[Wen et~al\mbox{.}(2018)]%
        {wen2018survey}
\bibfield{author}{\bibinfo{person}{Fei Wen}, \bibinfo{person}{Lei Chu}, \bibinfo{person}{Peilin Liu}, {and} \bibinfo{person}{Robert~C Qiu}.} \bibinfo{year}{2018}\natexlab{}.
\newblock \showarticletitle{A survey on nonconvex regularization-based sparse and low-rank recovery in signal processing, statistics, and machine learning}.
\newblock \bibinfo{journal}{\emph{IEEE Access}}  \bibinfo{volume}{6} (\bibinfo{year}{2018}), \bibinfo{pages}{69883--69906}.
\newblock


\bibitem[Wood(2003)]%
        {wood2003thin}
\bibfield{author}{\bibinfo{person}{Simon~N Wood}.} \bibinfo{year}{2003}\natexlab{}.
\newblock \showarticletitle{Thin plate regression splines}.
\newblock \bibinfo{journal}{\emph{Journal of the Royal Statistical Society: Series B (Statistical Methodology)}} \bibinfo{volume}{65}, \bibinfo{number}{1} (\bibinfo{year}{2003}), \bibinfo{pages}{95--114}.
\newblock


\bibitem[Wright et~al\mbox{.}(2010)]%
        {wright2010sparse}
\bibfield{author}{\bibinfo{person}{John Wright}, \bibinfo{person}{Yi Ma}, \bibinfo{person}{Julien Mairal}, \bibinfo{person}{Guillermo Sapiro}, \bibinfo{person}{Thomas~S Huang}, {and} \bibinfo{person}{Shuicheng Yan}.} \bibinfo{year}{2010}\natexlab{}.
\newblock \showarticletitle{Sparse representation for computer vision and pattern recognition}.
\newblock \bibinfo{journal}{\emph{Proc. IEEE}} \bibinfo{volume}{98}, \bibinfo{number}{6} (\bibinfo{year}{2010}), \bibinfo{pages}{1031--1044}.
\newblock


\bibitem[Zahid et~al\mbox{.}(2019)]%
        {zahid2019electricity}
\bibfield{author}{\bibinfo{person}{Maheen Zahid}, \bibinfo{person}{Fahad Ahmed}, \bibinfo{person}{Nadeem Javaid}, \bibinfo{person}{Raza~Abid Abbasi}, \bibinfo{person}{Hafiza~Syeda Zainab~Kazmi}, \bibinfo{person}{Atia Javaid}, \bibinfo{person}{Muhammad Bilal}, \bibinfo{person}{Mariam Akbar}, {and} \bibinfo{person}{Manzoor Ilahi}.} \bibinfo{year}{2019}\natexlab{}.
\newblock \showarticletitle{Electricity price and load forecasting using enhanced convolutional neural network and enhanced support vector regression in smart grids}.
\newblock \bibinfo{journal}{\emph{Electronics}} \bibinfo{volume}{8}, \bibinfo{number}{2} (\bibinfo{year}{2019}), \bibinfo{pages}{122}.
\newblock


\bibitem[Zhang et~al\mbox{.}(2015)]%
        {zhang2015survey}
\bibfield{author}{\bibinfo{person}{Zheng Zhang}, \bibinfo{person}{Yong Xu}, \bibinfo{person}{Jian Yang}, \bibinfo{person}{Xuelong Li}, {and} \bibinfo{person}{David Zhang}.} \bibinfo{year}{2015}\natexlab{}.
\newblock \showarticletitle{A survey of sparse representation: algorithms and applications}.
\newblock \bibinfo{journal}{\emph{IEEE access}}  \bibinfo{volume}{3} (\bibinfo{year}{2015}), \bibinfo{pages}{490--530}.
\newblock


\bibitem[Zhaoyang et~al\mbox{.}(2023)]%
        {eForecaster2023}
\bibfield{author}{\bibinfo{person}{Zhu Zhaoyang}, \bibinfo{person}{Chen Weiqi}, \bibinfo{person}{Xia Rui}, \bibinfo{person}{Zhou Tian}, \bibinfo{person}{Niu Peisong}, \bibinfo{person}{Peng Bingqing}, \bibinfo{person}{Wang Wenwei}, \bibinfo{person}{Liu Hengbo}, \bibinfo{person}{Ma Ziqing}, \bibinfo{person}{Wen Qingsong}, {and} \bibinfo{person}{Sun Liang}.} \bibinfo{year}{2023}\natexlab{}.
\newblock \showarticletitle{eForecaster: Unifying Electricity Forecasting with Robust, Flexible, and Explainable Machine Learning Algorithms}. In \bibinfo{booktitle}{\emph{Thirty-Seventh AAAI Conference on Artificial Intelligence}}. AAAI.
\newblock


\bibitem[Zhou et~al\mbox{.}(2022)]%
        {zhou2022fedformer}
\bibfield{author}{\bibinfo{person}{Tian Zhou}, \bibinfo{person}{Ziqing Ma}, \bibinfo{person}{Qingsong Wen}, \bibinfo{person}{Xue Wang}, \bibinfo{person}{Liang Sun}, {and} \bibinfo{person}{Rong Jin}.} \bibinfo{year}{2022}\natexlab{}.
\newblock \showarticletitle{Fedformer: Frequency enhanced decomposed transformer for long-term series forecasting}. In \bibinfo{booktitle}{\emph{International Conference on Machine Learning}}. PMLR, \bibinfo{pages}{27268--27286}.
\newblock


\end{thebibliography}

\begin{comment}

%%
%% If your work has an appendix, this is the place to put it.
\appendix

\section{Research Methods}

\subsection{Part One}

Lorem ipsum dolor sit amet, consectetur adipiscing elit. Morbi
malesuada, quam in pulvinar varius, metus nunc fermentum urna, id
sollicitudin purus odio sit amet enim. Aliquam ullamcorper eu ipsum
vel mollis. Curabitur quis dictum nisl. Phasellus vel semper risus, et
lacinia dolor. Integer ultricies commodo sem nec semper.

\subsection{Part Two}

Etiam commodo feugiat nisl pulvinar pellentesque. Etiam auctor sodales
ligula, non varius nibh pulvinar semper. Suspendisse nec lectus non
ipsum convallis congue hendrerit vitae sapien. Donec at laoreet
eros. Vivamus non purus placerat, scelerisque diam eu, cursus
ante. Etiam aliquam tortor auctor efficitur mattis.

\section{Online Resources}

Nam id fermentum dui. Suspendisse sagittis tortor a nulla mollis, in
pulvinar ex pretium. Sed interdum orci quis metus euismod, et sagittis
enim maximus. Vestibulum gravida massa ut felis suscipit
congue. Quisque mattis elit a risus ultrices commodo venenatis eget
dui. Etiam sagittis eleifend elementum.

Nam interdum magna at lectus dignissim, ac dignissim lorem
rhoncus. Maecenas eu arcu ac neque placerat aliquam. Nunc pulvinar
massa et mattis lacinia.

\end{comment}

\end{document}